\documentclass[preprint]{elsarticle}
\usepackage[title]{appendix}
\usepackage{algorithm}          % for algorithms header
\usepackage{algpseudocode}          % for algorithms descriptions
\usepackage[utf8]{inputenc}
\usepackage{graphicx}
\usepackage{float}
\usepackage{amsfonts}              % blackboard math symbols
\usepackage{amsmath}
\usepackage{multirow}              % table multirow functionality
\usepackage{subcaption}
\usepackage{wrapfig}
\usepackage[graphicx]{realboxes}
\usepackage{rotating}
\usepackage{diagbox}
\usepackage{array}
\usepackage{booktabs}
\usepackage{hyperref}
\usepackage{natbib}
\begin{document}

\begin{frontmatter}
% \title{Trainless Model Performance Estimation for Neural Architecture Search}
\title{Trainless Model Performance Estimation Based on Random Weights Initialisations for Neural Architecture Search}

\author[nims,tsukubauni]{Ekaterina Gracheva}
\ead{gracheva.ekaterina@nims.go.jp}
\address[nims]{International Center for Materials Nanoarchitectonics, National Institute for Materials Science, 1-1 Namiki, Tsukuba, Ibaraki, 305-0044 Japan}
\address[tsukubauni]{University of Tsukuba, 1-1-1 Tennodai, Tsukuba, Ibaraki, 305-8577 Japan}

\begin{abstract}
    Neural architecture search has become an indispensable part of the deep learning field. Modern methods allow to find one of the best performing architectures, or to build one from scratch, but they typically make decisions based on the trained accuracy information. In the present article we explore instead how the architectural component of a neural network affects its prediction power. We focus on relationships between the trained accuracy of an architecture and its accuracy prior to training, by considering statistics over multiple initialisations. We observe that minimising the coefficient of variation of the untrained accuracy, $CV_{U}$, consistently leads to better performing architectures. We test the $CV_{U}$ as a neural architecture search scoring metric using the NAS-Bench-201 database of trained neural architectures. The architectures with the lowest $CV_{U}$ value have on average an accuracy of $91.90 \pm 2.27$, $64.08 \pm 5.63$ and $38.76 \pm 6.62$ for CIFAR-10, CIFAR-100 and a downscaled version of ImageNet, respectively. Since these values are statistically above the random baseline, we make a conclusion that a good architecture should be stable against weights initialisations. It takes about $190$ s for CIFAR-10 and CIFAR-100 and $133.9$ s for ImageNet16-120 to process 100 architectures, on a batch of $256$ images, with $100$ initialisations.
\end{abstract}


\end{frontmatter}

\section{Introduction}

Since the beginning of the boom in the field of artificial intelligence, there has been continuous increase in data complexity and quantity, neural architecture designs, as well as yet increasing choice of powerful hardware. These factors render neural architecture building process complex. Given an extremely large number of parameters to be tuned, it can be extremely slow, when the decisions are made on a trial and error basis. Neural architecture search (NAS) is a way to automatise and accelerate the decision taking, shifting the task from humans to machines. It comes as no surprise, that recently NAS has become one of the most popular topics among the deep learning community.

The first attempts to find the most suitable network structure were done through evolutionary algorithms \cite{real2017large, liu2017hierarchical, suganuma2017genetic, elsken2018efficient}. There, several architectures are mutated in various ways (e.g. adding or removing a layer, changing activation function, etc.), and the resulting offsprings are evaluated through training. The best performing of the offsprings are added to the population for the next step, and the procedure is repeated for a given number of steps. This method has been used since back in the 1990's \cite{evolnas} and shows one of the best performances until now \cite{real2019regularized}.

Similarly, Bayesian optimisation \cite{frazier2018tutorial} is used to predict the best performing architecture out of many by training a subset of architectures \cite{shahriari2015taking}. This method has shown a few state-of-the-art performances in the period between 2013 and 2020 \cite{bergstra_bo1, domhan_bo2, mendoza2016towards_bo3, lambard2020}.

In 2016 Zoch et al. \cite{zoph2016neural} proposed to use the reinforcement learning to build neural architectures from scratch. There, a so-called controller neural network is trained to build a child-network — the network to be used for the final training and prediction. The original method demands tremendous amount of child-model training and is extremely lengthy. Several related works show significant acceleration of the process by reducing the search space \cite{zoph2018learning} or introducing weight sharing \cite{pham2018efficient}. An extensive overview of the NAS methods has been recently done by Thomas Elsken et al. \cite{elsken1808neural}.

The common point of all the above mentioned NAS algorithms is that at some point they all require model training. Not only that means longer search times, but also a higher uncertainty, since model training brings extra parameters to be tuned (\emph{e.g.}, batch size or learning rate).

As a step towards trainless NAS, in 2018 Istrate et al. \cite{istrate2019tapas} have introduced a small LSTM-based model, that allows to predict architecture's performance without training it on the data of interest. This model predicts an architecture's potential for a given data complexity. This data is taken from a so-called lifelong database of experiments. The straightaway restriction of this method is that there should already exist some data of a similar complexity within the database, and the available networks are limited to already existing ones (focused on image classification). Moreover, with time the overall procedure might lead to a bias, \textit{i.e.}, a most often predicted architecture in the beginning will have yet more chance to be output in future, thus "locking" it at the top position.

A similar approach is proposed by Deng et al. \cite{deng2017peephole}. They encode the layers composing the network into vectors, and bring them together with a predictive LSTM layer to build numerical representation of a network. A multilayer perceptron model is trained to predict the architecture with the highest prediction accuracy. Therefore, in order to use this method one needs to first train a set of architectures to acquire their trained accuracies, and then to train the predictive model on top. Since the final decision is made by a neural model, this method does not provide a reason why a given architecture has been chosen.

The first work that investigates fundamental architectural properties of neural networks in order to attain fully trainless NAS is proposed by Mellor et al. \cite{mellor2020neural} in 2020. The authors assess the neural architecture’s potential by passing a single minibatch of the data through a network forwards and backwards — one single time. Based on the results of the backpropagation, they measure the correlation between calculated gradients associated with the input layer. Using the NAS-Bench-201 benchmark database \cite{dong2020bench}, the authors show that their metric is able to distinguish one of the best neural architectures among many with consistent success. To the best of our knowledge, this is the only approach that aims to give an explanation of neural network's performance based on its structure.

On another end, there are a few papers, indicating that the best trained neural architecture often shows a better untrained accuracy. For example, the work of the UBER team \cite{zhou2019deconstructing} mentions that the best final architecture shows nearly $40\%$ accuracy on MNIST dataset \cite{lecun2010mnist} at initialisation. David Ha and Adam Gaier \cite{gaier2019weight} have presented a NAS algorithm which builds an architecture based on the untrained score. Their score is taking into account both the number of parameters contained within a model, which they seek to minimise, and the mean accuracy, which is being maximised. The mean accuracy is computed over several initialisations of the child model using a set of constant weights (single value for all the weights). They report that the resulting model achieves $82.0\% \pm 18.7\%$ on MNIST data with random weights at initialisation, and over $90\%$ when the weights are fixed to the best performing constant ones. 

These findings imply that neural networks might have an intrinsic property, which defines their prediction performance prior to training. Such property should not depend on the values of trainable parameters (weights), but only on network's topology. In order to cancel out the influence of the weights and to bring out the architectural component, we perform multiple random weights initialisations to assess averaged networks' performances. We compute several untrained statistics and explore their relationships with the trained accuracy. Based on the results of these tests, we deduce a trainless NAS scoring metric.

Our work can be divided into two parts. First, we have conducted an extensive MNIST study to explore dependencies between various untrained statistics and the trained accuracy. For this, we train a range of fully-connected neural networks on a reduced MNIST data, with multiple seeds and learning rates.\footnote{Fully reproducible code with data is available on GitHub at \url{https://github.com/egracheva/TrainlessNAS_MNIST}} Then, the most promising statistical property, the coefficient of variation $CV$, is tested on larger datasets and more complex neural geometries, to confirm its generality as a scoring metric for NAS.

The paper is structured as follows: Section \ref{par:methods} details the search spaces, datasets and training schemes used for the scoring metric search (\ref{subsec:mnistexplanation}) and application (\ref{subsec:nasbench201expl}). 
We present and discuss the results in Section \ref{par:results}. Subsection \ref{subsec:resultsigma} presents the selected scoring metric, while in Subsection \ref{subsec:resultnasbench201} we provide the results of the experiments with CIFAR-10, CIFAR-100 \cite{krizhevsky2009learning} and ImageNet16-120 \cite{chrabaszcz2017downsampled}. Conclusions and future improvements are proposed in Section \ref{par:conclusion}. 

\section{Materials and methods}
\label{par:methods}
\subsection{MNIST dataset processing and training}
\label{subsec:mnistexplanation}
\subsubsection{Dataset}

First, we explore correlations between some of the untrained performance statistics and the resulting trained accuracy evaluated on the test set. For this purpose we use a reduced version of MNIST \cite{lecun2010mnist} dataset, containing images of handwritten digits from 0 to 9. We reduce the size of the training set, leaving $20$ data points per class ($200$ data point in total). This is done to accelerate the training process and to train more models for better statistics. Besides, reduced training set makes the prediction task harder, which allows to distinguish the difference between architectures clearer. Note that both the validation and test sets are entirely preserved, containing $5000$ data points each. No data augmentation is applied. 

\subsubsection{Search space}
In order to reduce the uncertainty brought by complex neural structures (effects of initialisation, activation, etc.), the search region is limited to fully connected neural networks consisting of $2$ hidden layers. The number of units in each hidden layer is set to be one of the $12$ values in $[8, 16, 24, 32, 56, 64, 96, 128,$ $256, 512, 1024, 2048]$, making a total of $144$ of possible architectures.

\subsubsection{Training scheme}
Every neural network is initialised and trained with $100$ different seeds between $0$ and $99$, and $6$ learning rates $[0.0001, 0.0003, 0.001, 0.003, 0.01, 0.03]$ ($600$ trainings per architecture, $86400$ trainings overall). The batch size $N_{BS}$ is fixed to $50$, which we found showing the best results for a wide range of architectures within the search space. The models are built with Keras \cite{chollet2015keras} and Tensorflow \cite{tensorflow2015-whitepaper} and trained for $200$ epochs using $3$ NVIDIA Titan V GPUs. Weights are initialised using the He uniform initialiser \cite{he2015delving}, which is used together with ReLU activation function \cite{nair2010rectified} for hidden layers and Adam optimiser \cite{adam} with default decay rates ($0.9$ and $0.99$ for the first and second moments, respectively). 

The final weights are based on the epoch with the best validation accuracy after a burn-in period of $50$ epochs. Ignoring the first quarter of the training process is based on experience, since the validation loss of small noisy data tends to demonstrate random behaviour in the beginning of the training, leading to faulty results.

The pseudocode for the MNIST \cite{lecun2010mnist} training process is given in Algorithm \ref{algo1}.

\begin{algorithm}
\caption{MNIST training pseudocode}
\begin{algorithmic}
\State Load the data
\State Split data on train/val/test sets
\For{cat in categories}
    \Comment creating reduced train set
    \State Randomly pick $20$ points from the original train set
\EndFor
\For{nunits\_layer\_1 in $[8, 16, \dots, 2048]$}
    \For{nunits\_layer\_2 in $[8, 16, \dots, 2048]$}
        \For{lr in $[0.0001, 0.0003, 0.001, 0.003, 0.01, 0.03]$}
            \For{seed in range($N_{init}$)}
                \State Build initial model
                \State Assess the untrained accuracy $U_{i}$
                \Comment untrained accuracy
                \State Train the model
                \State Select the final weights based on the best validation accuracy
                \State Compute test set accuracy $T_{i}$
                \Comment trained accuracy
            \EndFor
            \State Compute means and standard deviations
            \begin{gather*}
            \mu_{T}=\frac{1}{N}\sum\limits_{i=1}^{N_{init}}T_{i}, \quad
            \sigma_{T}=\sqrt{\frac{\sum\limits_{i=1}^{N_{init}}(T_{i}-\mu_{T})^2}{N_{init}}}\\
            \mu_{U}=\frac{1}{N}\sum\limits_{i=1}^{N_{init}}U_{i}, \quad
            \sigma_{U}=\sqrt{\frac{\sum\limits_{i=1}^{N_{init}}(U_{i}-\mu_{U})^2}{N_{init}}}
            \end{gather*}
        \EndFor
        \State Select the best performing learning rate based on $max(\mu_{T})$
        \State
        \Comment one set of [$\mu_{T}$, $\sigma_{T}$, $\mu_{U}$, $\sigma_{U}$] per architecture
    \EndFor
\EndFor
\end{algorithmic}
\label{algo1}
\end{algorithm}

Once the training is complete, only the learning rate showing the highest average training accuracy is selected for each architecture. This is done to insure that neural architectures are compared in a fair way, each showing its best performance. Afterwards, mean untrained error $\mu_{U}$, mean trained error $\mu_{T}$, together with their respective standard deviations ($\sigma_{U}$, $\sigma_{T}$) are calculated.

\subsection{CIFAR-10, CIFAR-100 and ImageNet}
\label{subsec:nasbench201expl}
\subsubsection{Search space: NAS-Bench-201}
To test more complex geometries on challenging datasets, we used a modified version of the code used by Mellor et al. \cite{mellor2020neural}, published together with their paper\footnote{The code can be found on GitHub at \url{https://github.com/BayesWatch/nas-without-training}}. To check the validity of their NAS search metric, the authors use the NAS-Bench-201 search space \cite{dong2020bench}. It is a set of architectures with a fixed skeleton, consisting of convolution layer and three stacks of cells, connected by a residual block. Each cell is a densely-connected directed acyclic graph with $4$ nodes, $5$ possible operations and no limits on the number of edges, providing a total of $15,625$ possible architectures.
\subsubsection{Datasets}
Each of the architectures from NAS-Bench-201 \cite{dong2020bench} is trained on three major datasets: CIFAR-10, CIFAR-100 \cite{krizhevsky2009learning} and ImageNet \cite{chrabaszcz2017downsampled}. 
Since the original CIFAR datasets do not contain a validation set, the NAS-Bench-201 authors created one by splitting the original data. In case of CIFAR-10, the training set is split into halves to form the validation set, leaving the test set unchanged; for CIFAR-100, the test set is split in halves to form the validation set and the new test set. For the sake of computational tractability, a simplified version of ImageNet is used \cite{chrabaszcz2017downsampled}. All the images are down-scaled to 16x16 pixels, with 120 classes kept, forming a new ImageNet16-120 dataset. Data augmentation is used for all datasets; augmentation schemes differ slightly between CIFAR \cite{krizhevsky2009learning} and ImageNet \cite{chrabaszcz2017downsampled} due to the difference between input image sizes.

An overview on all the data used in the present work is given in Table \ref{tab:data}.

\begin{table*}[]
\caption{A summary over the datasets used in this paper: number of classes, image resolution and splitting schemes (in thousands) for reduced MNIST \cite{lecun2010mnist}, CIFAR-10, CIFAR-100 \cite{krizhevsky2009learning} and ImageNet16-120 \cite{chrabaszcz2017downsampled}.}
\centering
\setlength{\tabcolsep}{3pt}
\begin{tabular*}{\textwidth}{@{\extracolsep{\fill}}llll}
\hline
Dataset         & Classes & Resolution & Train/val/test (K) \\
\hline
Reduced MNIST   & 10      & 28x28      & 0.2/5/5            \\
CIFAR-10        & 10      & 32x32x3    & 25/25/10           \\
CIFAR-100       & 100     & 32x32x3    & 50/5/5             \\
ImageNet16-120  & 120     & 16x16x3    & 151.7/3/3          \\
\hline
\end{tabular*}
\label{tab:data}
\end{table*}

\subsubsection{Training}
The training is done using up to $3$ different seeds, and with the same fixed set of hyperparameters for each dataset. The authors use stochastic gradient descent with Nesterov momentum, batch size $N_{BS}=256$, learning rate between $0.1$ and $0$ with cosine annealing and weight decay of $5 \times 10^{-4}$. Architectures are trained for $200$ epochs.

\subsubsection{Experimental scheme}
The goal of this part of the study is to determine how efficiently does a given scoring metric select a good architecture among many random ones. In order to obtain statistically significant information, selection process is run $N_{runs}=500$ times, each time choosing $N_{a}$ architectures at random (among $15,625$ available). Each architecture is initialised $N_{init}$ times, in order to access the mean $\mu_{U}$ and standard deviation $\sigma_{U}$ of the untrained performance. The batch data used for the accuracy computation is fixed for every individual run, so that all the architectures are fairly compared, and there is no uncertainty coming from the data choice. The pseudocode for this part of the study is given in Algorithm \ref{algo2}. For this, we use a modified code provided by Mellow
et al. together with their paper \cite{mellor2020neural}.\footnote{The modified code can be found on GitHub at \url{https://github.com/egracheva/TrainlessNAS_NAS201Bench}}

\begin{algorithm}[H]
\caption{$CV_{U}$ tests on NAS-Bench-201}
\begin{algorithmic}
\For{run in range($N_{runs}$)}
    \State Randomly select $N_{BS}$ images from the training dataset
    \State Randomly select $N_{a}$ architectures from the whole space
    \Comment arches
    \For{arch in arches}
        \For{seed in range($N_{init}$)}
            \State Initialise the arch with the seed
            \State Forward propagate selected $N_{BS}$ images
            \State Compute untrained accuracy $U_{i}$
        \EndFor
        \State Compute mean $\mu_{U}$, standard deviation $\sigma_{U}$ for untrained accuracies over initialisations
        \begin{gather*}
        \mu_{U}=\frac{1}{N}\sum\limits_{i=1}^{N_{init}}U_{i}, \quad
        \sigma_{U}=\sqrt{\frac{\sum\limits_{i=1}^{N_{init}}(U_{i}-\mu_{U})^2}{N_{init}}}
        \end{gather*}
        \State Compute the score
        \begin{equation}
            CV_{U} = \frac{\sigma_{U}}{\mu_{U}}
        \end{equation}
    \EndFor
    \State Select the architecture with the minimum score value ($CV_{U}>0$)
    \State Retrieve trained accuracy $T$ for the selected architecture from the database
\EndFor
\State Average trained accuracies of selected architectures over $N_{runs}$
\begin{equation}
    \mu_{T} = \sum\limits_{j=1}^{N_{runs}}T_{j}
\end{equation}
\end{algorithmic}
\label{algo2}
\end{algorithm}

Filtering out the scores equal to zero is necessary for the random architectures containing no meaningful layers (for example, architectures consisting of skip-connection layers only). These architectures, naturally, show random accuracy with no deviation ($\sigma_{U} = 0$).

\section{Results and discussion}
\label{par:results}
\subsection{Scoring metric search with MNIST}
\label{subsec:resultsigma}
\begin{figure}[H]
  \begin{subfigure}{0.48\textwidth}
    \includegraphics[width=\linewidth]{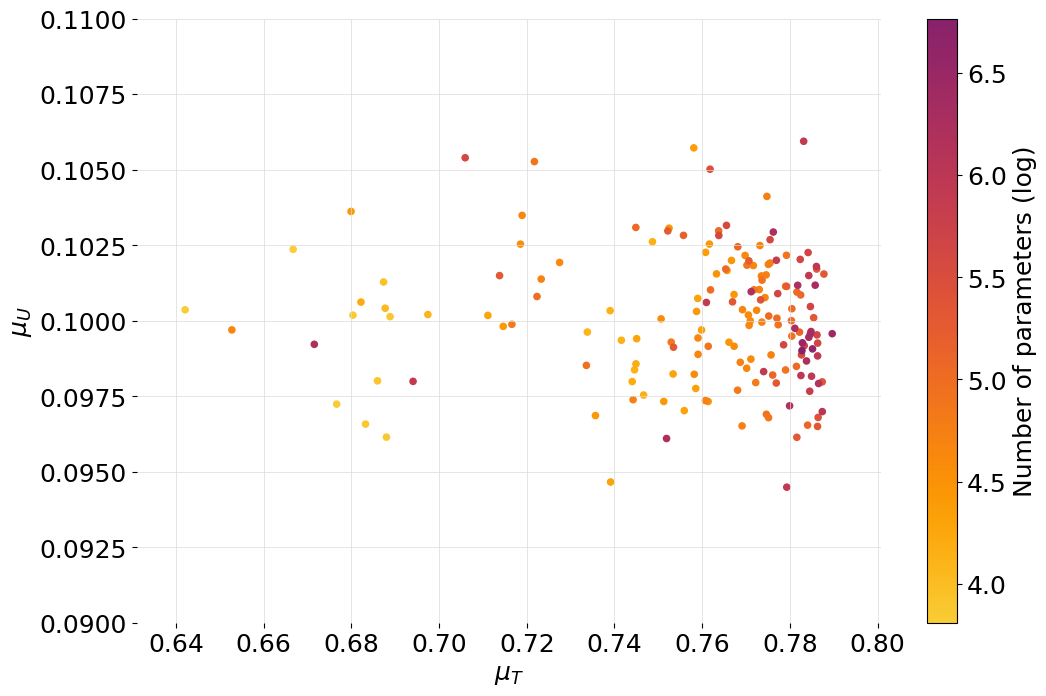}
    \caption{ } \label{fig:1a}
  \end{subfigure}
  \hspace*{\fill}   % maximise separation between the subfigures
  \begin{subfigure}{0.48\textwidth}
    \includegraphics[width=\linewidth]{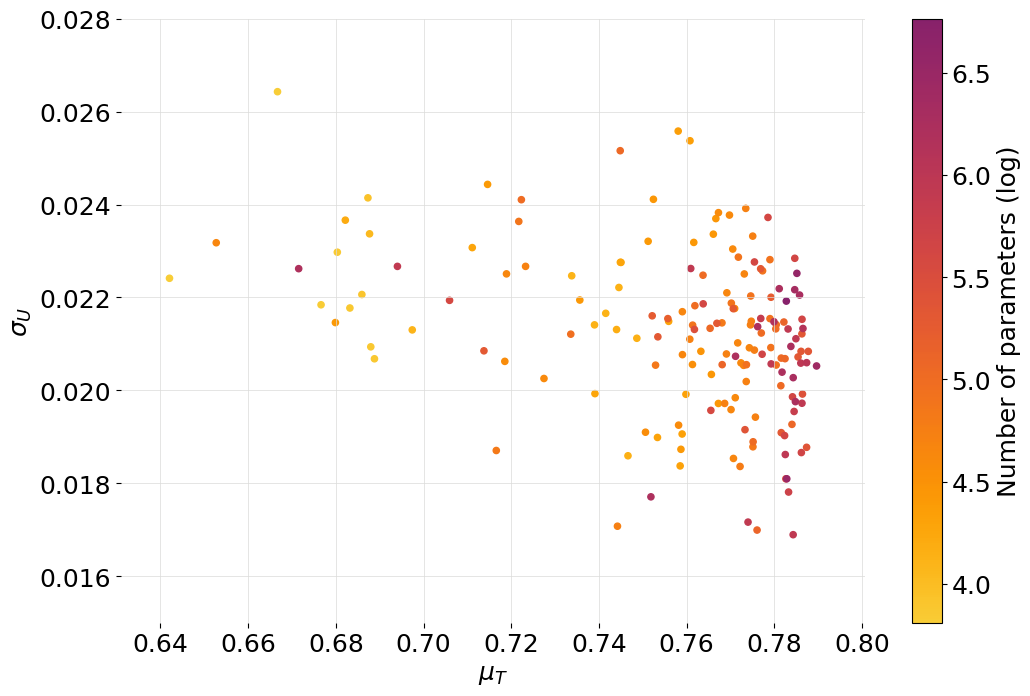}
    \caption{ } \label{fig:1b}
  \end{subfigure}
\caption{(a) Mean untrained accuracy $\mu_{U}$ (a) and standard deviation of untrained accuracy $\sigma_{U}$ (b) against mean trained accuracy $\mu_{T}$, all three computed over $N_{init}=100$ initialisations. One point stands for one architecture. The colours represent the logarithm of the number of parameters for a given architecture.} 
\label{fig:dependancies}
\end{figure}

The aim of the experiments related to MNIST \cite{lecun2010mnist} is to explore dependencies between various untrained statistics and the trained accuracy. The existing machine learning literature suggests that the best trained architecture may also show high untrained performance \cite{gaier2019weight, zhou2019deconstructing}. We expect, thus, to see some tendency between mean accuracies prior to and after the training. We denote these accuracies as $\mu_{U}$ and $\mu_{T}$, respectively. Against our expectations, there is no clear correlation between these two metrics, as shown in Figure \ref{fig:1a}. Instead, surprisingly, the mean trained accuracy $\mu_{T}$ seems to be related to the untrained standard deviation $\sigma_{U}$: even though there is no linear correlation, the lowest $\sigma_{U}$ values belong to architectures from the top performance range (Figure \ref{fig:1b}).

\begin{figure}[H]
    \centering
    \begin{minipage}{0.48\textwidth}
        \centering
        \includegraphics[width=\textwidth]{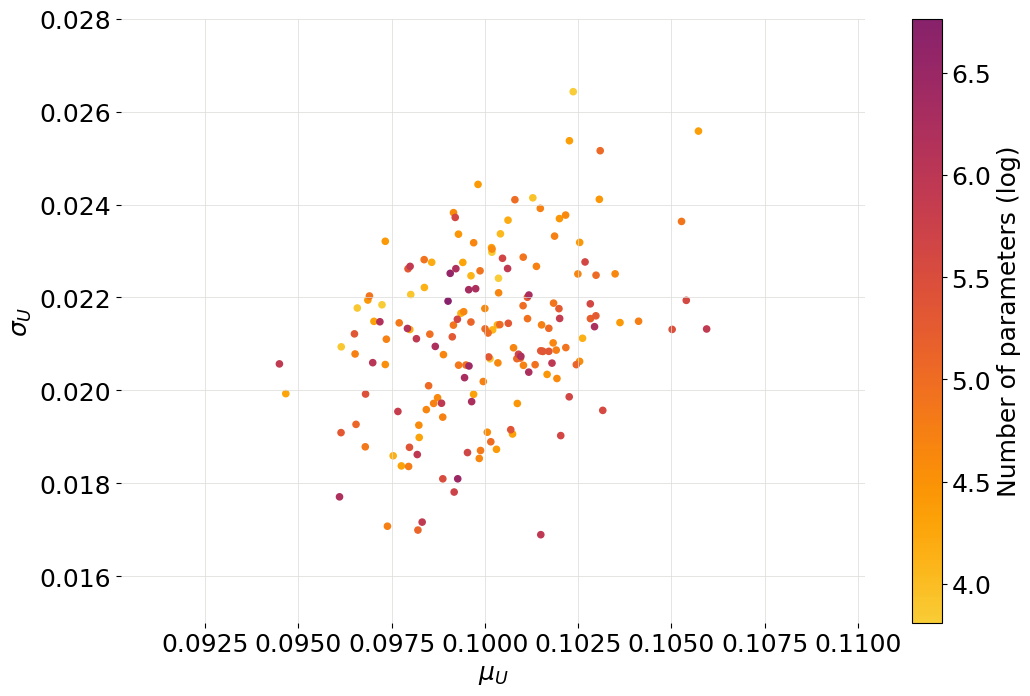}
        \caption{Mean untrained accuracy $\mu_{U}$ against standard deviation of untrained accuracy $\sigma_{U}$, computed over $N_{init}=100$ initialisations. One point stands for one architecture. The colours represent the logarithm of the number of parameters for a given architecture.}
        \label{fig:musigmacorr}
    \end{minipage}\hfill
    \begin{minipage}{0.48\textwidth}
        \centering
        \includegraphics[width=\textwidth]{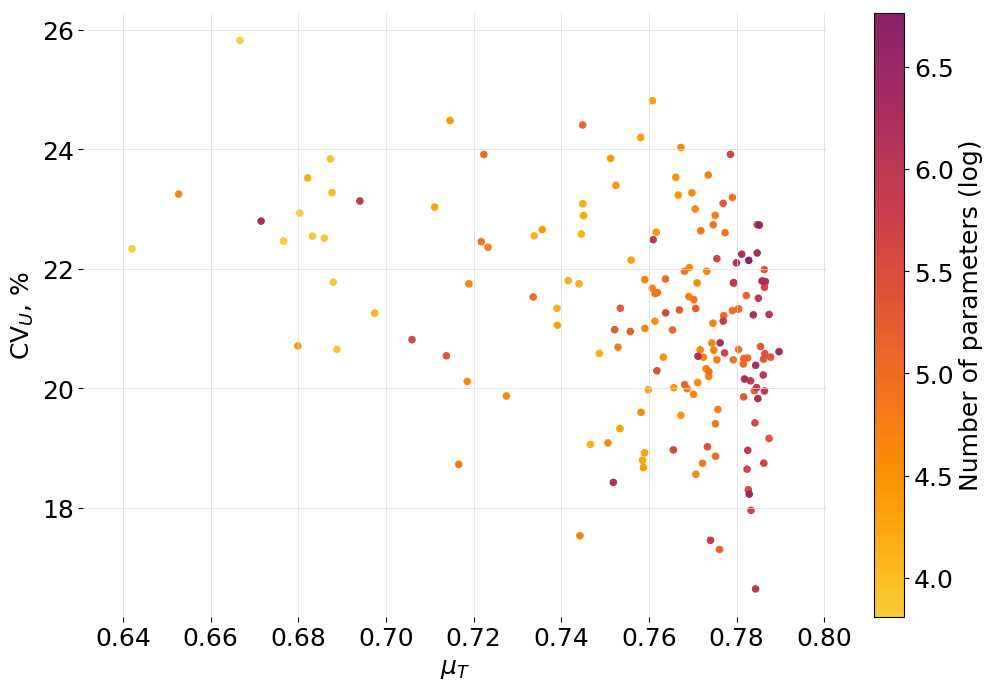} 
        \caption{Coefficient of variation of the untrained accuracy $CV_{U}$ ($\rm\%$) against mean trained accuracy $\mu_{T}$, both computed over $N_{init}=100$ initialisations. One point stands for one architecture. The colours represent the number of parameters contained within an architecture.}
        \label{fig:mutsigmar}
    \end{minipage}
\end{figure}

We have also observed that lower means $\mu_{U}$ corresponds to lower standard deviations $\sigma_{U}$ (Figure \ref{fig:musigmacorr}). Indeed, lower accuracy values lead to proportionally lower mean and standard deviation. Therefore, minimising standard deviation alone may bias towards the networks that show overall low untrained accuracies $U_{i}$. To compensate for this effect, we normalise the standard deviation $\sigma_{U}$ by the mean $\mu_{U}$:
\begin{gather*}
CV_{U} = \frac{\sigma_{U}}{\mu_{U}}.
\end{gather*}

The resulting parameter $CV_{U}$ is known in statistics as the coefficient of variation, or relative standard deviation. When plotting the coefficient of variation $CV_{U}$ against the trained accuracy $\mu_{T}$ in Figure \ref{fig:mutsigmar}, tendency becomes yet more clear: selecting the architectures with low $CV_{U}$ leads to high trained accuracy $\mu_{T}$.

\begin{figure}[H]
  \begin{subfigure}{0.48\textwidth}
    \includegraphics[width=\linewidth]{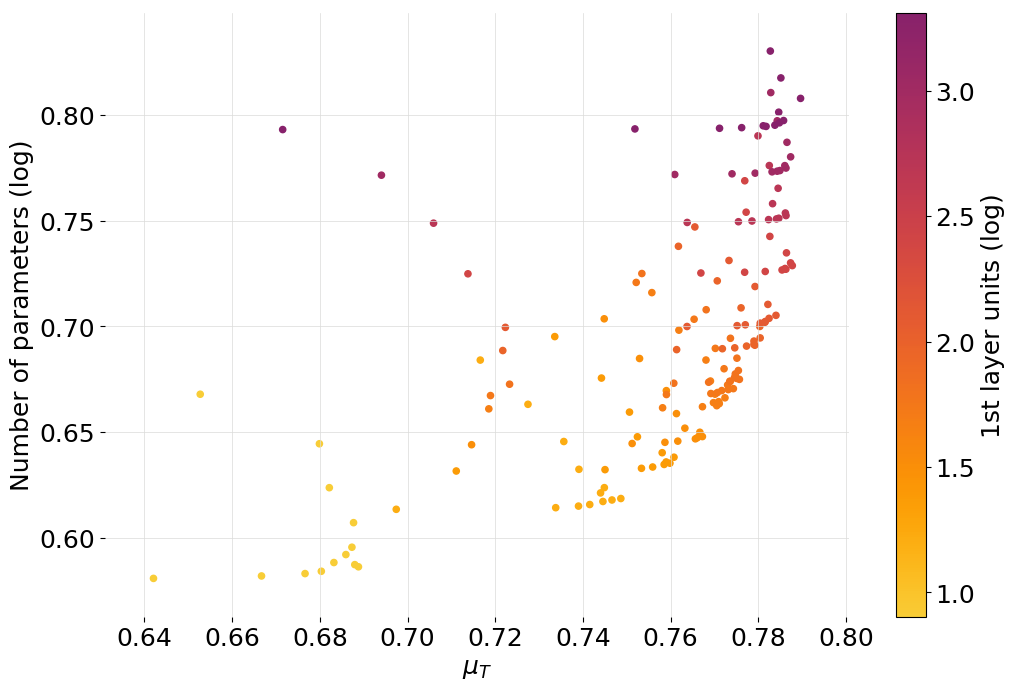}
    \caption{ } \label{fig:4a}
  \end{subfigure}
  \begin{subfigure}{0.48\textwidth}
    \includegraphics[width=\linewidth]{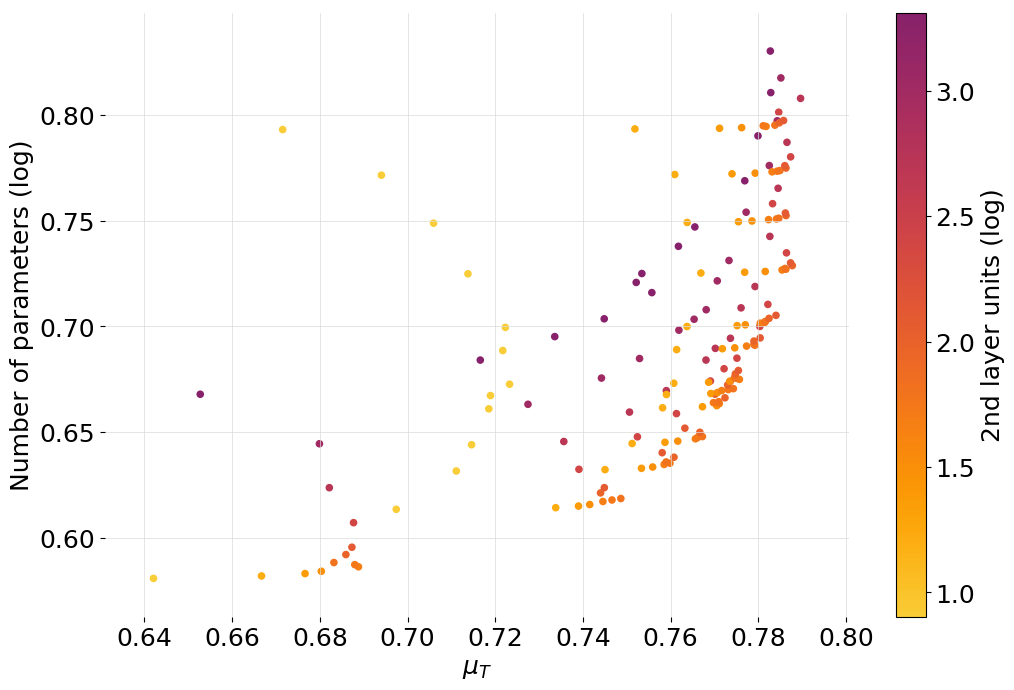}
    \caption{ } \label{fig:4b}
  \end{subfigure}
\caption{Number of parameters against the architectures mean trained performance $\mu_{T}$, computed over $N_{init}=100$ initialisations. One point represents one architecture. The colours represent the number of units in the first layer (a) and the second layer (b).}
\label{fig:mutnuints}
\end{figure}

When choosing a NAS scoring metric, one has to consider how it correlates with the number of parameters contained within the network. It has been shown earlier that bigger does not necessarily mean better \cite{dauphin2013big, lecun1990optimal}. Even though there is a higher chance for a bigger network to contain a subnetwork, capable of successfully fitting the data \cite{frankle2018lottery}, there is also an increasing risk of overfitting, and increasing training time. We can confirm the effect of the performance saturation with our toy MNIST model both for the totality of parameters, and for the parameters in a single layer, as demonstrated in Figure \ref{fig:mutnuints}.

\begin{wrapfigure}{R}{0.48\textwidth}
    \centering
    \includegraphics[width=0.48\textwidth]{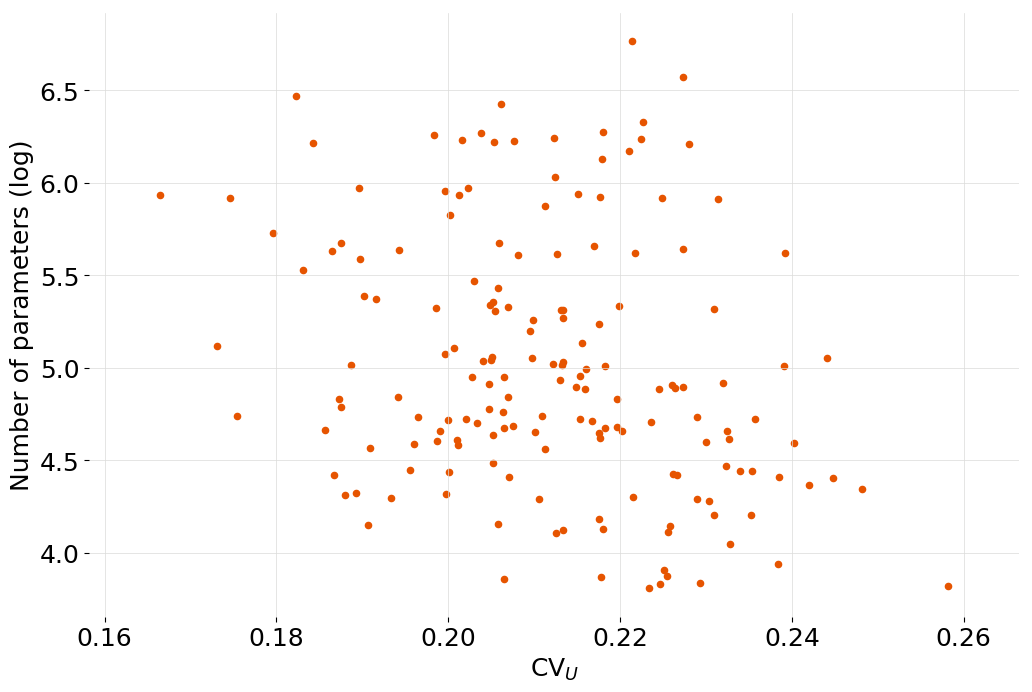}
    \caption{The number of parameters against the scoring metric $CV_{U}$ ($\rm\%$), or the coefficient of variation of the untrained performance, computed over $N_{init}=100$ initialisations. One point stands for one architecture.} 
    \label{fig:nparamnunits}
\end{wrapfigure}
 Therefore, in order to find optimal architecture regardless the number of parameters, one should use a scoring metric uncorrelated with them. Figure \ref{fig:nparamnunits} shows that there is no significant correlation between $CV_{U}$ and the number of parameters.

Taking all the above into consideration, we conclude that $CV_{U}$ is a suitable scoring metric for NAS.

\subsection{Testing the scoring metric \texorpdfstring{$CV_{U}$}{Lg}}
\label{subsec:resultnasbench201}
The results of the $CV_{U}$ performance with CIFAR-10, CIFAR-100 \cite{krizhevsky2009learning} and ImageNet16-120 \cite{chrabaszcz2017downsampled} are given in Table \ref{tab:nasbench201table}. We present our results based on $100$ initialisations ($N_{init}=100$), for $N_{BS}=256$, both used by Mellor et al. \cite{mellor2020neural} and during the NAS-Bench-201 training. We also provide the two sample t-test p-values for the statistical significance of differences between our results and those of Mellor et al. \cite{mellor2020neural}, as well as for the random baseline (p-value $<0.05$ means that the results are statistically different, otherwise, they are considered similar). Comparisons are made both with the best performing $N_{a}$ and with $N_{a}$ fixed to $100$ architectures. The results show that the performance of the $CV_{U}$ scoring metric is clearly above random for all three datasets. 

\begin{table}[]
\centering
\small
\setlength{\tabcolsep}{3pt}
\renewcommand{\arraystretch}{1.2}
\resizebox{\textwidth}{!} {
\begin{tabular}{llllp{0.01\textwidth}llp{0.01\textwidth}ll}
\hline
\multirow{2}{*}{Method}  &\multirow{2}{*}{Time(s)} & \multicolumn{2}{l}{CIFAR-10} & \multirow{2}{*}{} & \multicolumn{2}{l}{CIFAR-100} & \multirow{2}{*}{} & \multicolumn{2}{l}{ImageNet16-120}  \\
\cline{3-4}\cline{6-7}\cline{9-10}
 &   & validation & test &   & validation & test &   & validation & test\\
\hline
\multicolumn{10}{c}{State-of-the-art}\\
REA       & $12000$ & $91.19 \pm 0.31$  & $93.92 \pm 0.3$  &   & $71.81 \pm 1.12$  & $71.84 \pm 0.99$ &   & $45.15 \pm 0.89$ & $45.54 \pm 1.03$ \\
Random Search        & $12000$ & $90.93 \pm 0.36$  & $93.92 \pm 0.31$ &   & $70.93 \pm 1.09$  & $71.04 \pm 1.07$ &   & $44.45 \pm 1.1$  & $44.57 \pm 1.25$ \\
REINFORCE & $12000$ & $91.09 \pm 0.37$  & $93.92 \pm 0.32$ &   & $71.61 \pm 1.12$  & $71.71 \pm 1.09$ &   & $45.05 \pm 1.02$ & $45.24 \pm 1.18$ \\
BOHB      & $12000$ & $90.82 \pm 0.53$  & $93.92 \pm 0.33$ &   & $70.74 \pm 1.29$  & $70.85 \pm 1.28$ &   & $44.26 \pm 1.36$ & $44.42 \pm 1.49$ \\
\hline
\multicolumn{10}{c}{Baselines}\\
Optimal ($N_{a}=10$)  & N/A & $89.92 \pm 0.75$ & $93.06 \pm 0.59$ &   & $69.61 \pm 1.21$ & $69.76 \pm 1.25$ &   & $43.11 \pm 1.85$ & $43.30 \pm 1.87$  \\
Optimal ($N_{a}=100$) & N/A & $91.05 \pm 0.28$ & $93.84 \pm 0.23$ &   & $71.45 \pm 0.79$ & $71.56 \pm 0.78$ &   & $45.37 \pm 0.61$ & $45.67 \pm 0.64$  \\
Random & N/A & $83.20 \pm 13.28$ & $86.61 \pm 13.46$ &   & $60.70 \pm 12.55$ & $60.83 \pm 12.58$ &   & $33.34 \pm 9.39$ & $33.13 \pm 9.66$ \\
\hline
\multicolumn{10}{c}{Trainless}\\
Mellor et al. ($N_{a}=10$) & $1.7$  & \pmb{$88.47 \pm 1.3$}   & $91.53 \pm 1.62$ &   & $66.49 \pm 3.08$  & $66.63 \pm 3.14$ &   & \pmb{$38.33 \pm 4.98$} & \pmb{$38.33 \pm 5.22$} \\
Mellor et al. ($N_{a}=25$) & $4.8$  & $88.46 \pm 1.42$   & \pmb{$91.78 \pm 1.45$} &   & \pmb{$66.87 \pm 2.84$}  & \pmb{$67.05 \pm 2.89$} &   & $37.18 \pm 6.11$ & $37.07 \pm 6.39$          \\
Mellor et al. ($N_{a}=100$)  & $17.4$ & $88.45 \pm 1.46$   & $91.61 \pm 1.71$ &   & $66.42 \pm 3.27$  & $66.56 \pm 3.28$ &   & $36.56 \pm 6.7$  & $36.37 \pm 6.97$          \\
\noalign{\vskip 1.5mm} 
 $CV_{U}$ ($N_{a}=10$)   & $19.6/19.5/14.9$  & \pmb{$85.01 \pm 6.08$} & $91.03 \pm 2.77$ &   & $63.73 \pm 5.62$ & $63.83 \pm 5.65$ &   & $37.74 \pm 6.89$ & $37.70 \pm 7.13$  \\
 $CV_{U}$ ($N_{a}=25$)   &  $47.7/47.5/34.0$ & $84.91 \pm 5.98$ & $91.46 \pm 2.39$ &   & $63.83 \pm 5.50$ & $63.92 \pm 5.57$ &   & $38.42 \pm 6.30$ & $38.42 \pm 6.51$  \\
 $CV_{U}$ ($N_{a}=100$)   & $190.0/190.4/133.9$  & $84.89 \pm 6.39$ & \pmb{$91.90 \pm 2.27$} &   & \pmb{$63.99 \pm 5.61$} & \pmb{$64.08\pm 5.63$} &   & \pmb{$38.68 \pm 6.34$} & \pmb{$38.76 \pm 6.62$}  \\
\hline
\noalign{\vskip 1mm}
p-values (best $N_{a}$) &   &   & \pmb{$0.319$} &   &   & \pmb{$3.7\mathrm{e}{-24}$} &   &   & \pmb{$0.254$} \\
p-values ($N_{a}=100)$  &   &   & $0.023$ &   &   & $8\mathrm{e}{-17}$             &   &   & $3\mathrm{e}{-8}$ \\
\hline
p-values (random baseline) &   &   & \pmb{$5\mathrm{e}{-11}$} &   &   & \pmb{$2\mathrm{e}{-7}$} &   &   & \pmb{$2\mathrm{e}{-25}$} \\
\end{tabular}
}
\caption{Comparison of the trainless $CV_{U}$ metric performance against existing NAS algorithms on CIFAR-10, CIFAR-100 \cite{krizhevsky2009learning} and ImageNet16-120 \cite{chrabaszcz2017downsampled} datasets. On the top, we list the best performing methods that require training (REA \cite{real2019regularized}, random search, REINFORCE \cite{williams1992simple}, BOHB \cite{falkner2018bohb}). As a low limit reference, the random and optimal values for $N_{a}\in\{10, 100\}$ are given. Then, the results from Mellor et al. \cite{mellor2020neural} and our results are reported for $N_{a}\in\{10, 25, 100\}$ with $N_{BS}=256$. Our training elapsed times are reported in CIFAR-10/CIFAR-100/ImageNet16-120 format. Finally, the two sample t-test p-values are provided for two cases: when comparing best performing $N_{a}$ (bold), and with a fixed $N_{a}$=100.}
\label{tab:nasbench201table}
\end{table}

\begin{figure}
    \centering
    \begin{minipage}{.3\linewidth}
        \begin{subfigure}{\linewidth}
            \includegraphics[width=\textwidth]{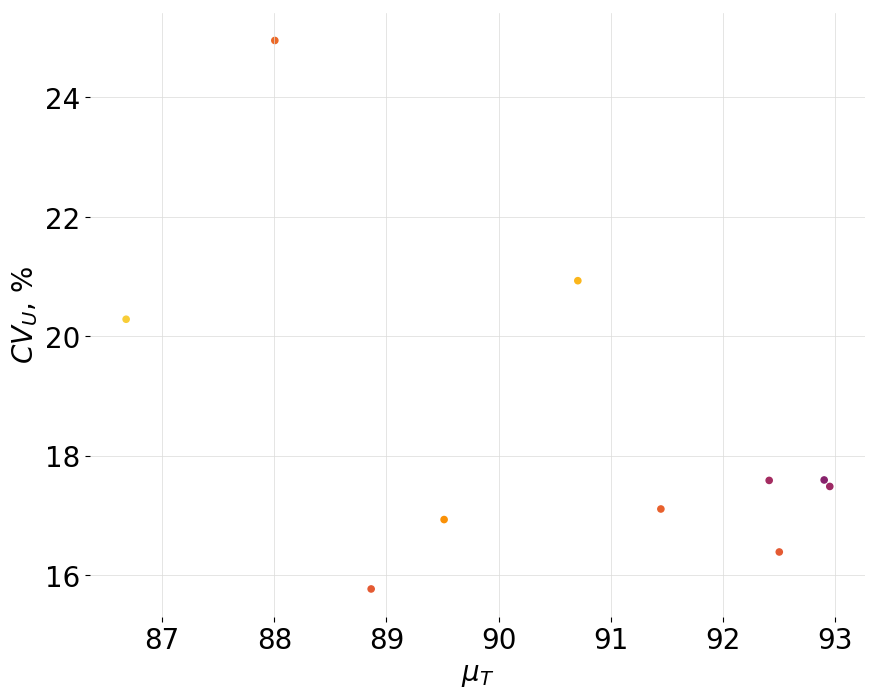}
            \caption{$N_{a}=10$}
        \end{subfigure} \\
        \begin{subfigure}{\linewidth}
            \includegraphics[width=\textwidth]{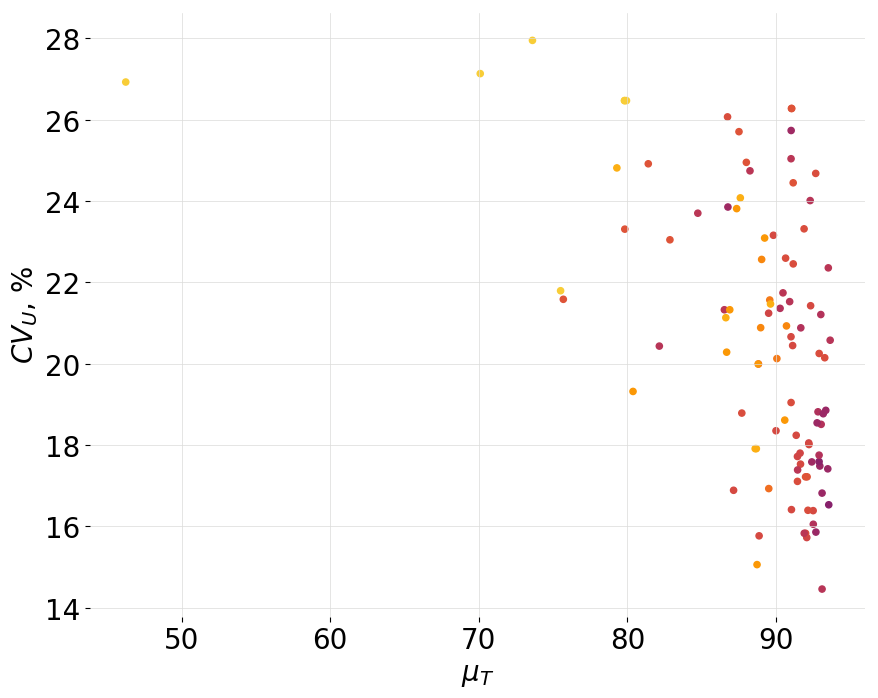}
            \caption{$N_{a}=100$}
        \end{subfigure} 
    \end{minipage}
    \begin{minipage}{.3\linewidth}
        \begin{subfigure}{\linewidth}
            \includegraphics[width=\textwidth]{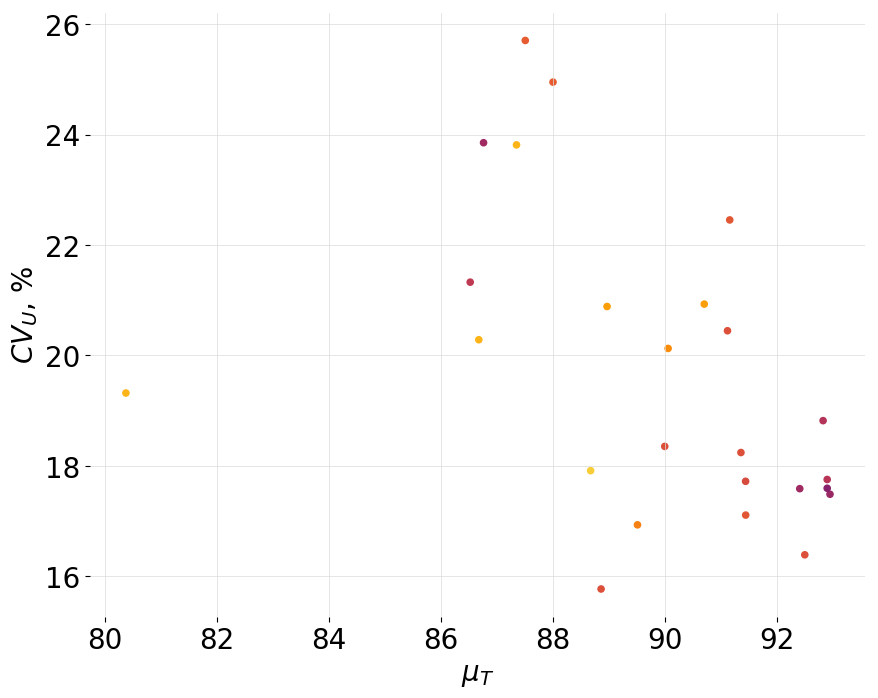}
            \caption{$N_{a}=25$}
        \end{subfigure} \\
        \begin{subfigure}{\linewidth}
            \includegraphics[width=\textwidth]{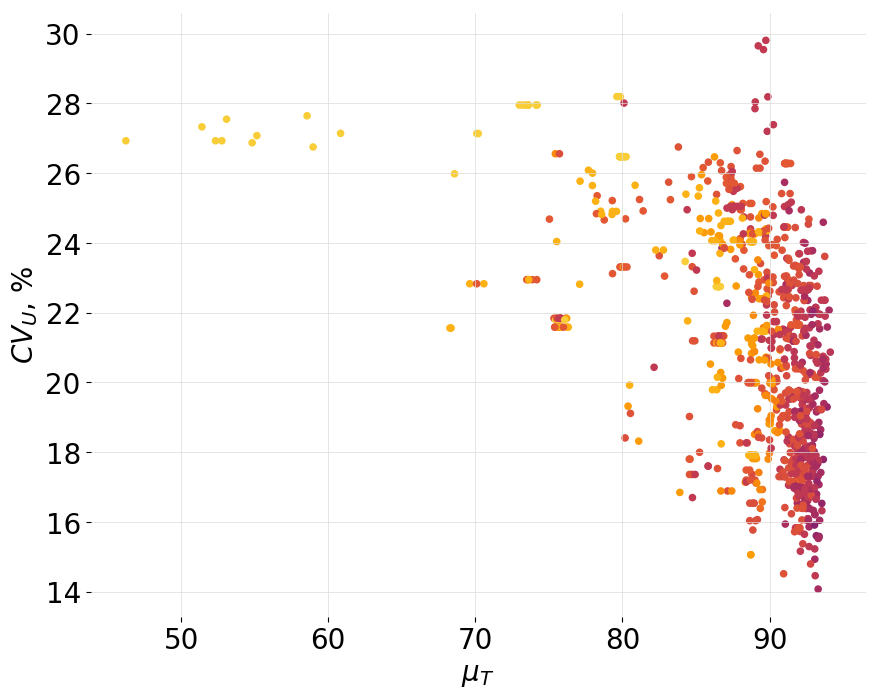}
            \caption{$N_{a}=1000$}
        \end{subfigure} 
    \end{minipage}
    \begin{minipage}{.3\linewidth}
        \begin{subfigure}{\linewidth}
            \includegraphics[width=\textwidth]{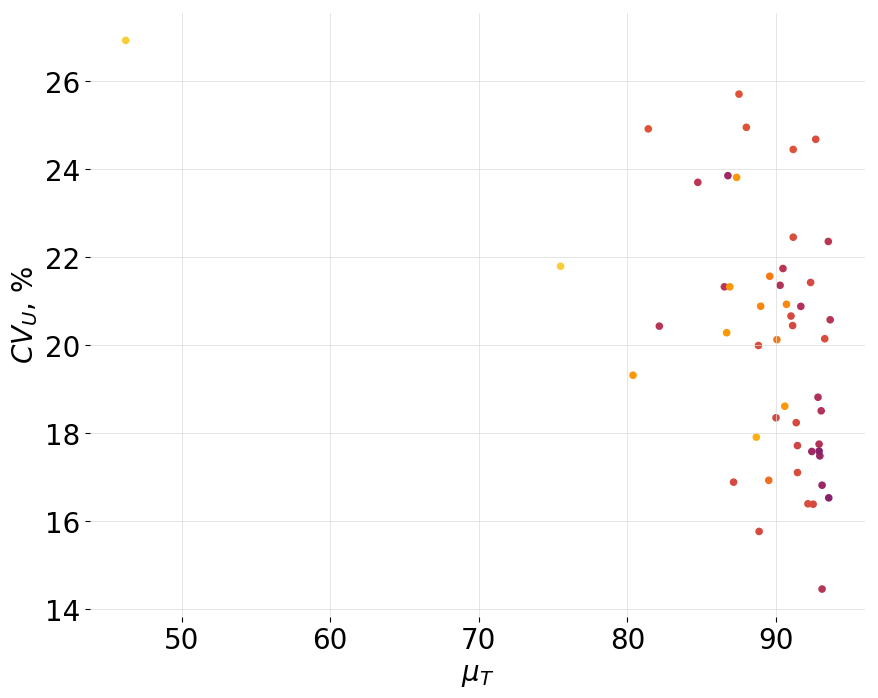}
            \caption{$N_{a}=50$}
        \end{subfigure} \\
        \begin{subfigure}{\linewidth}
            \includegraphics[width=\textwidth]{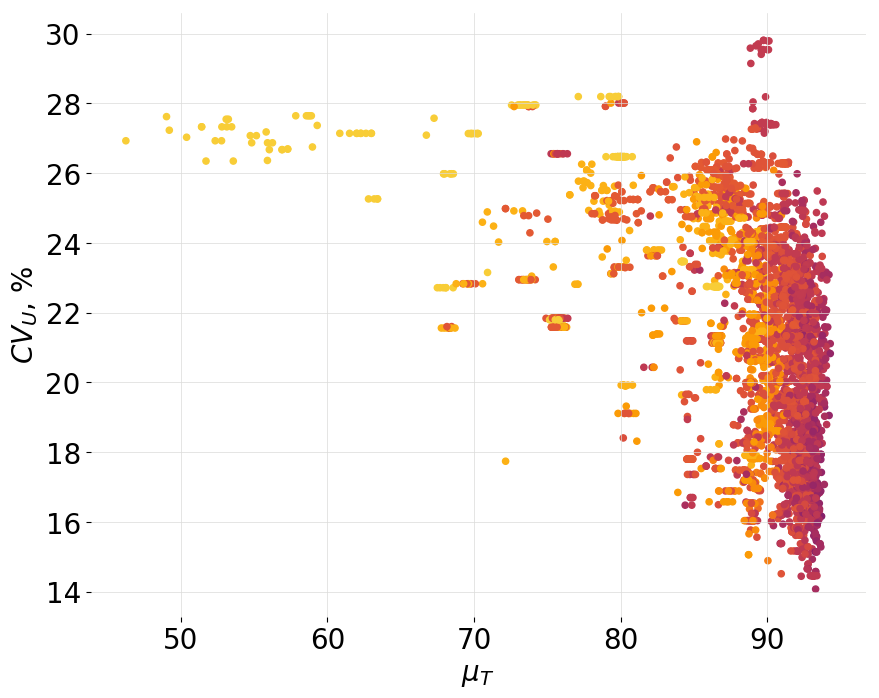}
            \caption{$N_{a}=5000$}
        \end{subfigure} 
    \end{minipage}
    \begin{minipage}{.07\linewidth}
            \begin{subfigure}{\linewidth}
                \includegraphics[width=\textwidth]{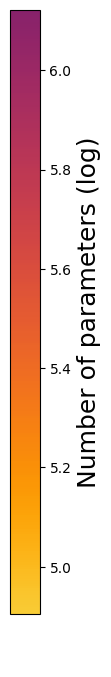}
            \end{subfigure}
        \end{minipage}
\caption{Comparison of the coefficient of variation $CV_{U}$ performance against mean trained accuracy $\mu_{T}$ for CIFAR-10 \cite{krizhevsky2009learning} dataset for different number of selected architectures $N_{a}\in[10, 25, 50, 100, 1000, 5000]$. Statistics are computed over $N_{init}=100$ initialisations. One point stands for one architecture. The colours represent the logarithm of the total number of trainable parameters.} 
\label{fig:cifar10arch}
\end{figure}

\begin{figure}
    \centering
    \begin{minipage}{.3\linewidth}
        \begin{subfigure}{\linewidth}
            \includegraphics[width=\textwidth]{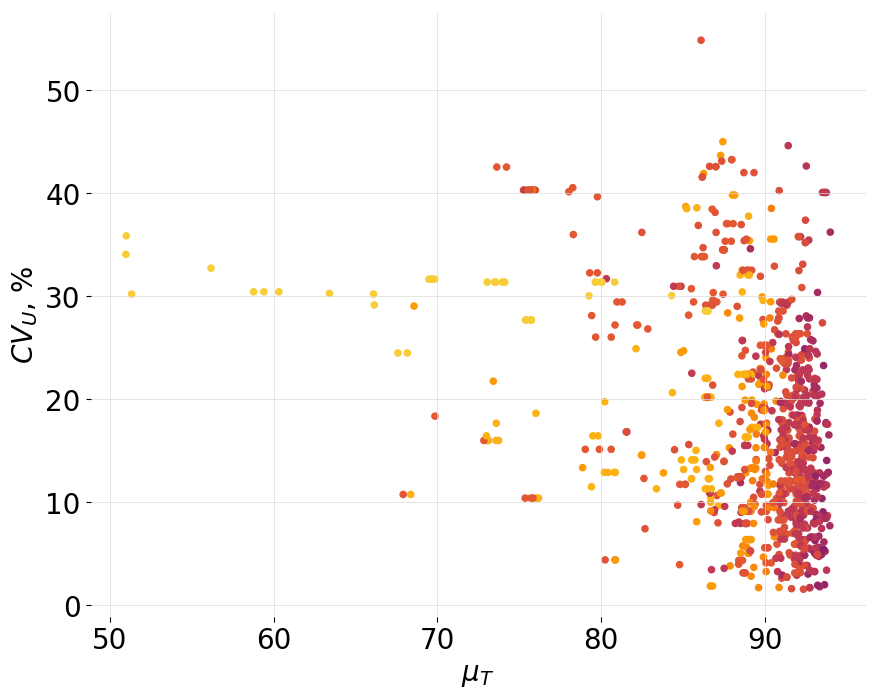}
            \caption{$N_{init}=3$}
        \end{subfigure} \\
        \begin{subfigure}{\linewidth}
            \includegraphics[width=\textwidth]{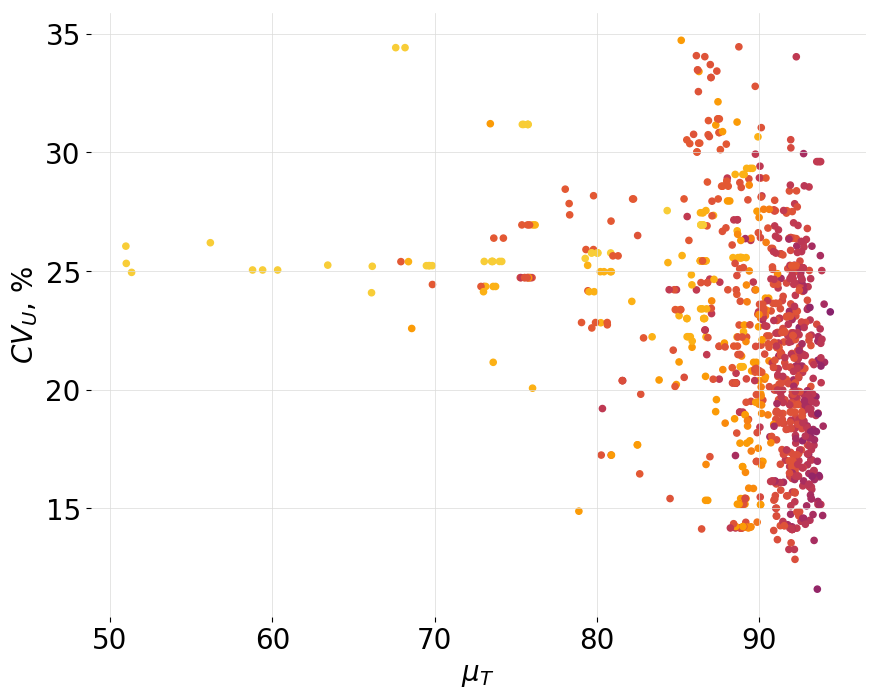}
            \caption{$N_{init}=25$}
        \end{subfigure} 
    \end{minipage}
    \begin{minipage}{.3\linewidth}
        \begin{subfigure}{\linewidth}
            \includegraphics[width=\textwidth]{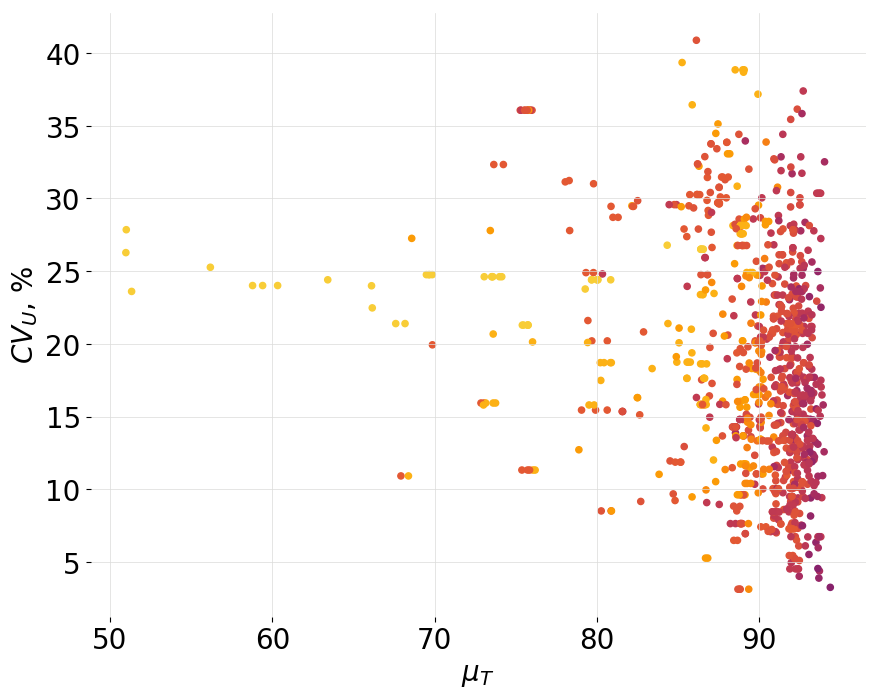}
            \caption{$N_{init}=5$}
        \end{subfigure} \\
        \begin{subfigure}{\linewidth}
            \includegraphics[width=\textwidth]{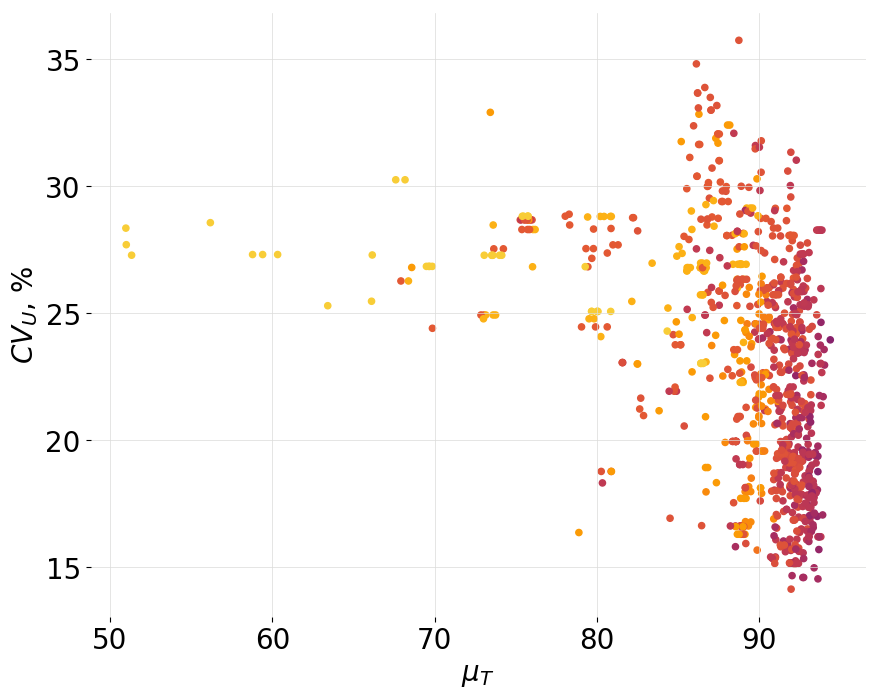}
            \caption{$N_{init}=50$}
        \end{subfigure} 
    \end{minipage}
    \begin{minipage}{.3\linewidth}
        \begin{subfigure}{\linewidth}
            \includegraphics[width=\textwidth]{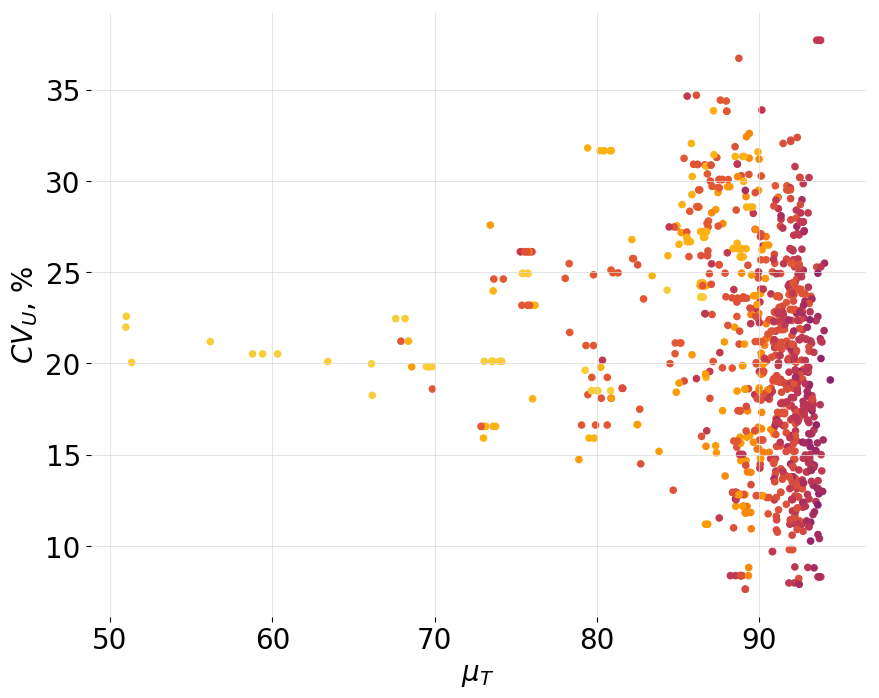}
            \caption{$N_{init}=10$}
        \end{subfigure} \\
        \begin{subfigure}{\linewidth}
            \includegraphics[width=\textwidth]{figures/cifar10/Correlaction_Score_vs_Accuracy_CIFAR10_256bs_100it_1000arch.png}
            \caption{$N_{init}=100$}
        \end{subfigure} 
    \end{minipage}
    \begin{minipage}{.07\linewidth}
            \begin{subfigure}{\linewidth}
                \includegraphics[width=\textwidth]{figures/ColorBar.png}
            \end{subfigure}
        \end{minipage}
\caption{Comparison of the coefficient of variation $CV_{U}$ performance against mean trained accuracy $\mu_{T}$ for CIFAR-10 \cite{krizhevsky2009learning} dataset. Statistics are computed over varying number of initialisations $N_{init}\in[3, 5, 10, 25, 50, 100]$. Number of architectures $N_{a}=1000$. One point stands for one architecture. Colours represent the logarithm of the total number of trainable parameters.} 
\label{fig:cifar10init}
\end{figure}

The effects of number of iterations and number of selected architectures are shown in Figures \ref{fig:cifar10arch} and \ref{fig:cifar10init}, respectively, on an example of CIFAR-10 \cite{krizhevsky2009learning}. The number of picked architectures considerably increases the overall performance, since there is more chance to involve a good architecture. The number of iterations improves the precision of the method. Similar plots for CIFAR-100 \cite{krizhevsky2009learning} and ImageNet16-120 \cite{chrabaszcz2017downsampled} can be found in Appendix (Figures \ref{fig:cifar100arch}, \ref{fig:cifar100init}, \ref{fig:imagenet16120arch}, \ref{fig:imagenet16120init}). Table \ref{tab:overalltab} shows results of our metric performance with various batch sizes, $N_{init}$ and $N_{a}$ combinations.

%CIFAR10, architectures

% CIFAR10, initialisations

We compare our results against the results presented by Mellor et al., since in their work they also aim to discover a direct architectural property. We do not make comparison with other trainless NAS methods, since they rely on a supplementary model responsible for the architecture choice. In Table \ref{tab:nasbench201table}, similar overall performances are observed. Methods are also similar in the sense that they filter out bad architectures, rather than choose the best one. 

Mellor et al. focus on correlations between linear maps (Jacobians) of input entries. Jacobian of a given input expresses how much local perturbations within this input impact the corresponding output. Their metric minimises the correlation between Jacobians within a minibatch using the eigenvalues of the correlation matrix:

\begin{gather*}
S = -\sum\limits_{i=1}^n [\log(\sigma_{J,i} + k) + (\sigma_{J,i} + k)^{-1}],
\end{gather*}

where $\sigma_{J,i}$ are the eigenvalues of $\Sigma$, and k is a small constant added for numerical stability ($k = 1\mathrm{e}{-5}$). It's worth noting that the choice of the final score's shape is not clearly explained in \cite{mellor2020neural}. 

The success of the $S$ metric means, that when inputs affect the output in an uncorrelated way, the neural network has a higher chance to distinguish between them, and therefore to have a better trainability. Their method, however, depends slightly on the values of the initial weights.
 
Our approach, on the other hand, focuses on how much variation in weights affects the outputs. $CV_{U}$ quantifies the stability of the network against initialisations for the same fixed data minibatch. Intuitively, if a network is stable against random weights, it will also be less affected by weights fluctuations during the training. It might suggest that the function representing a stable network is relatively smooth, which allows for more efficient training and lower overfitting risks.

As it was mentioned above, our algorithm involves two extra hyperparameters, which may be considered as a disadvantage. The first one is the batch size: there are significant deviations on the prediction power (with different optimal $N_{BS}$ for each dataset, see Table \ref{tab:overalltab}). The second is the number of initialisations. Besides, the fact that our method requires multiple initialisations leads to a significantly slower performance compared to Mellor et al. (running time grows linearly with the number of initialisations). Yet, comparing to the methods that require training, the absolute performance speed remains high (tens to hundreds of seconds).

Prediction accuracy improves with the number of sampled architectures (for any batch size). This is a natural consequence of the fact that the chance of having a well performing architecture among many architectures is higher than among few (which is confirmed by random selection tests, see Table \ref{tab:nasbench201table}). Note that in the work of Mellor et al. \cite{mellor2020neural} increasing the number of sampled architectures does not improve the result, which is counterintuitive. While this could be a statistical artefact for CIFAR \cite{krizhevsky2009learning} data, for ImageNet \cite{chrabaszcz2017downsampled} the difference between $N_{a}=10$ and $N_{a}=100$ is statistically significant (p-value of $5.8\mathrm{e}{-7}$).

Nevertheless, the $CV_{U}$ metric alone is not sufficient for successful NAS. It can be partly justified by the fact that all the architectures within the NAS-201-Benchmark are trained with the same fixed set of hyperparameters. For some of the networks contained within the benchmark these set may not be optimal. There is a possibility that the architectures selected by our metric could have achieved better accuracies. We plan to investigate it in future work, as well as to try to combine our metric with other NAS methods (for example, the one from Mellor et al. \cite{mellor2020neural}).

\section{Conclusions}
\label{par:conclusion}
In this work we explore relashionship between the prediction performance of an architecture and its accuracy prior to training. The principal objective is to better understand how the neural network's geometry affects its prediction power. For this, we evaluate untrained accuracy over multiple random weights initialisations. We observe that the architectures with low coefficient of variation of untrained accuracy $CV_{U} = \sigma_{U} / \mu_{U}$ show overall better performance. 
We use this observation to develop an entirely trainless NAS technique. Our metric achieves the accuracies of $91.90 \pm 2.27$, $64.08 \pm 5.63$ and $38.76 \pm 6.62$ for CIFAR-10, CIFAR- \cite{krizhevsky2009learning} and a downscaled version of ImageNet \cite{chrabaszcz2017downsampled}, respectively (when choosing among $100$ architectures, with $100$ random initialisations and evaluating accuracies on a minibatch of $256$ data points). These accuracies are statistically above the random baseline, which leads to the conclusion that the stability of a network against initialisations is an indicator of its trainability. However, since this metric does not guarantee the best architecture at all times, we consider that the stability is not the only property that influences the neural architecture's performance. Combining our method with others (for example, the one from Mellor et al. \cite{mellor2020neural}), might lead to more stable results. We plan to explore various combinations of the $CV_{U}$ metric with other methods in future work.

\section{Acknowledgement}
We would like to express our deepest gratitude to Dr. Ayako Nakata and Dr. Guillaume Lambard for their continuous support and valuable discussions.

\bibliography{references}

\clearpage
\appendix
\section{Appendix}
\begin{table}[h]
\caption{}
\small
\setlength{\tabcolsep}{3pt}
\newcolumntype{H}{>{\setbox0=\hbox\bgroup}c<{\egroup}@{}}
\resizebox{\textwidth}{!}{
\begin{tabular}{@{\extracolsep{4pt}}llllllllllll}
\toprule
\noalign{\vskip 1.5mm}
\multicolumn{12}{c}{CIFAR-100}\\
\noalign{\vskip 1.5mm} 
% \cline{3-20}
\toprule
\noalign{\vskip 1.5mm} 
\multicolumn{12}{c}{Batch size}\\
\cline{3-4}\cline{5-6}\cline{7-8}\cline{9-10}\cline{11-12}
\multicolumn{2}{l}{ } & \multicolumn{2}{c}{2} & \multicolumn{2}{c}{4}           & \multicolumn{2}{c}{8}           & \multicolumn{2}{c}{16}          & \multicolumn{2}{c}{32}           \\
  $N_{a}$     &    $N_{init}$  & Validation      & Test            & Validation     & Test           & Validation     & Test           & Validation     & Test           & Validation     & Test                     \\
\hline
\multirow{2}{*}{10}  & $10$         & $59.44 \pm 12.50$ & $59.54 \pm 12.51$ & $65.71 \pm 6.03$ & $65.86 \pm 6.02$ & $62.71 \pm 8.44$ & $62.84 \pm 8.44$ & $62.00 \pm 8.75$ & $62.09 \pm 8.82$ & $62.46 \pm 8.16$ & $62.58 \pm 8.19$ \\
                    & $100$        & $62.30 \pm 9.05$ & $62.44 \pm 9.07$ & $62.62 \pm 7.28$ & $62.71 \pm 7.27$ & $62.33 \pm 7.58$ & $62.44 \pm 7.57$ & $63.65 \pm 6.93$ & $63.71 \pm 6.99$ & $59.60 \pm 8.95$ & $59.71 \pm 9.04$ \\
\hline
\multirow{2}{*}{25} & $10$        & $58.00 \pm 13.99$ & $58.11 \pm 13.95$ & $66.33 \pm 4.84$ & $66.48 \pm 4.84$ & $65.67 \pm 5.22$ & $65.79 \pm 5.22$ & $64.60 \pm 5.23$ & $64.67 \pm 5.30$ & $60.45 \pm 9.01$ & $60.52 \pm 9.08$ \\
                    & $100$         & $60.14 \pm 10.26$ & $60.23 \pm 10.24$ & $62.38 \pm 7.46$ & $62.48 \pm 7.49$ & $62.47 \pm 5.97$ & $62.57 \pm 6.00$ & $63.76 \pm 6.84$ & $63.84 \pm 6.91$ & $58.77 \pm 9.31$ & $58.86 \pm 9.37$ \\
\hline
\multirow{2}{*}{100} & $10$         & $58.06 \pm 13.65$ & $58.19 \pm 13.64$ & $66.64 \pm 3.24$ & $66.74 \pm 3.27$ & $66.20 \pm 4.02$ & $66.27 \pm 4.05$ & $64.80 \pm 4.85$ & $64.94 \pm 4.94$ & $59.60 \pm 8.95$ & $59.71 \pm 9.04$ \\
                    & $100       $ & $60.42 \pm 8.19$ & $60.49 \pm 8.18$ & $63.30 \pm 6.11$ & $63.38 \pm 6.09$ & $61.87 \pm 5.83$ & $61.98 \pm 5.87$ & $65.78 \pm 5.29$ & $65.88 \pm 5.32$ & $60.08 \pm 8.62$ & $60.15 \pm 8.71$\\
                    
\noalign{\vskip 3mm}
\multicolumn{12}{c}{Batch size}\\
\cline{3-4}\cline{5-6}\cline{7-8}\cline{9-10}
\multicolumn{2}{l}{ } & \multicolumn{2}{c}{64} & \multicolumn{2}{c}{128}           & \multicolumn{2}{c}{256}           & \multicolumn{2}{c}{512}   &        \\
  $N_{a}$     &    $N_{init}$  & Validation      & Test            & Validation     & Test           & Validation     & Test           & Validation     & Test & & \\
\cline{1-2} \cline{3-10}
\multirow{2}{*}{10}  & $10$         & $62.20 \pm 7.88 $ & $62.33 \pm 7.89 $ & $63.21 \pm 7.11$ & $63.31 \pm 7.19$ & $62.96 \pm 7.68$ & $63.05 \pm 7.67$ & $62.20 \pm 7.97$ & $62.32 \pm 8.00$ & \\
                     & $100$        & $59.77 \pm 10.35$ & $59.88 \pm 10.38$ & $60.46 \pm 8.83$ & $60.59 \pm 8.86$ & $63.73 \pm 5.62$ & $63.83 \pm 5.65$ & $64.28 \pm 5.45$ & $64.38 \pm 5.49$  & \\
\cline{1-2} \cline{3-10}
\multirow{2}{*}{25}  & $10$         & $62.49 \pm 7.37 $ & $62.61 \pm 7.40 $ & $63.42 \pm 7.02$ & $63.53 \pm 7.08$ & $61.91 \pm 8.13$ & $62.05 \pm 8.14$ & $61.51 \pm 8.48$ & $61.63 \pm 8.51$ & \\
                     & $100$        & $60.03 \pm 9.68 $ & $60.15 \pm 9.68 $ & $60.34 \pm 7.97$ & $60.45 \pm 8.00$ & $63.83 \pm 5.50$ & $63.92 \pm 5.57$ & $64.15 \pm 5.11$ & $64.23 \pm 5.19$  & \\
\cline{1-2} \cline{3-10}
\multirow{2}{*}{100} & $10$         & $63.10 \pm 6.03 $ & $63.21 \pm 6.01 $ & $63.74 \pm 5.93$ & $63.89 \pm 5.97$ & $61.98 \pm 7.44$ & $62.09 \pm 7.45$ & $60.82 \pm 8.81$ & $60.96 \pm 8.82$  & \\
                     & $100$        & $60.54 \pm 8.54 $ & $60.64 \pm 8.56 $ & $60.91 \pm 7.84$ & $61.01 \pm 7.87$ & $63.99 \pm 5.61$ & $64.08 \pm 5.63$ & $64.01 \pm 5.06$ & $64.10 \pm 5.10$   & \\
\noalign{\vskip 10mm}
\toprule
\noalign{\vskip 1.5mm}
\multicolumn{12}{c}{CIFAR-10}\\
\noalign{\vskip 1.5mm}
\toprule
\noalign{\vskip 1.5mm} 
\multicolumn{12}{c}{Batch size}\\
\cline{3-4}\cline{5-6}\cline{7-8}\cline{9-10}\cline{11-12}

\multicolumn{2}{l}{ } & \multicolumn{2}{c}{2} & \multicolumn{2}{c}{4}           & \multicolumn{2}{c}{8}           & \multicolumn{2}{c}{16}          & \multicolumn{2}{c}{32}         \\
% \cline{3-4}\cline{5-6}\cline{7-8}\cline{9-10}\cline{11-12}
  $N_{a}$     &    $N_{init}$  & Validation      & Test            & Validation     & Test           & Validation     & Test           & Validation     & Test           & Validation     & Test    \\
 \cline{1-2} \cline{3-12}
\multirow{2}{*}{10}  & $10$         & $84.02 \pm 7.41$ & $88.45 \pm 5.70$ & $85.90 \pm 4.04$ & $89.03 \pm 4.75$ & $85.28 \pm 5.78$ & $90.79 \pm 3.21$ & $86.62 \pm 4.00$ & $88.67 \pm 7.46$ & $83.20 \pm 7.18$ & $87.52 \pm 8.04$ \\
                    & $100$        & $83.94 \pm 6.10$ & $90.68 \pm 2.94$ & $85.79 \pm 5.33$ & $91.50 \pm 2.71$ & $83.32 \pm 8.00$ & $91.46 \pm 2.42$ & $85.01 \pm 6.08$ & $90.96 \pm 3.07$ & $85.63 \pm 5.05$ & $90.55 \pm 3.24$\\
 \cline{1-2} \cline{3-12}
\multirow{2}{*}{25} & $10$        & $84.14 \pm 6.68$ & $88.42 \pm 4.93$ & $85.98 \pm 4.11$ & $89.37 \pm 4.16$ & $85.42 \pm 6.16$ & $91.08 \pm 3.38$ & $86.49 \pm 4.15$ & $89.89 \pm 5.49$ & $83.04 \pm 6.37$ & $87.62 \pm 7.68$ \\
                    & $100$         & $88.42 \pm 4.93$ & $90.79 \pm 3.51$ & $89.37 \pm 4.16$ & $91.82 \pm 2.55$ & $91.08 \pm 3.38$ & $91.56 \pm 1.74$ & $89.89 \pm 5.49$ & $90.89 \pm 2.96$ & $87.62 \pm 7.68$ & $91.15 \pm 2.29$ \\
 \cline{1-2} \cline{3-12}
\multirow{2}{*}{100} & $10$         & $84.80 \pm 4.52$ & $87.99 \pm 5.02$ & $85.51 \pm 3.72$ & $89.85 \pm 3.55$ & $86.23 \pm 5.05$ & $91.57 \pm 1.77$ & $86.37 \pm 3.59$ & $90.49 \pm 4.09$ & $82.02 \pm 6.49$ & $87.99 \pm 8.17$\\
                    & $100       $ & $83.81 \pm 5.20$ & $91.04 \pm 2.50$ & $87.04 \pm 3.99$ & $92.37 \pm 1.85$ & $82.68 \pm 7.62$ & $91.57 \pm 1.65$ & $84.89 \pm 6.39$ & $91.43 \pm 1.64$ & $86.23 \pm 4.13$ & $91.36 \pm 1.96$\\
                    
\noalign{\vskip 3mm}
\multicolumn{12}{c}{Batch size}\\
\cline{3-4}\cline{5-6}\cline{7-8}\cline{9-10}
\multicolumn{2}{l}{ } & \multicolumn{2}{c}{64}            & \multicolumn{2}{c}{128}         & \multicolumn{2}{c}{256}         & \multicolumn{2}{c}{512} & \\  
$N_{a}$     &    $N_{init}$ & Validation      & Test            & Validation     & Test           & Validation     & Test           & Validation     & Test \\
\cline{1-2} \cline{3-10}

\multirow{2}{*}{10}  & $10$         & $84.65 \pm 6.66 $ & $87.31 \pm 7.86 $ & $85.31 \pm 5.42$ & $88.65 \pm 7.35$ & $85.60 \pm 5.02$ & $89.97 \pm 3.89$ & $86.35 \pm 4.25$ & $91.15 \pm 2.48$  & \\
                     & $100$        & $86.19 \pm 4.94 $ & $91.19 \pm 2.71 $ & $87.59 \pm 2.55$ & $90.49 \pm 3.67$ & $87.98 \pm 1.95$ & $91.03 \pm 2.77$ & $88.29 \pm 1.64$ & $91.48 \pm 1.82$  & \\
\cline{1-2} \cline{3-10}
\multirow{2}{*}{25}  & $10$         & $83.82 \pm 6.81 $ & $87.31 \pm 7.54 $ & $85.71 \pm 4.72$ & $87.98 \pm 8.87$ & $85.48 \pm 5.36$ & $90.13 \pm 3.59$ & $86.68 \pm 2.95$ & $91.34 \pm 1.93$ & \\
                     & $100$        & $86.29 \pm 3.60 $ & $91.75 \pm 2.17 $ & $87.63 \pm 2.21$ & $91.14 \pm 3.22$ & $88.03 \pm 1.74$ & $91.46 \pm 2.39$ & $88.27 \pm 1.48$ & $91.51 \pm 1.75$  & \\
\cline{1-2} \cline{3-10}
\multirow{2}{*}{100} & $10$         & $83.80 \pm 6.39 $ & $88.18 \pm 7.12 $ & $86.27 \pm 3.97$ & $88.88 \pm 7.90$ & $86.05 \pm 4.56$ & $90.33 \pm 3.86$ & $86.44 \pm 2.57$ & $91.25 \pm 2.23$  & \\
                     & $100$        & $86.39 \pm 3.31 $ & $92.50 \pm 1.59 $ & $87.49 \pm 2.46$ & $92.32 \pm 2.16$ & $88.18 \pm 1.66$ & $91.90 \pm 2.27$ & $88.39 \pm 1.37$ & $91.52 \pm 1.87$ & \\  

\noalign{\vskip 10mm}
\toprule
\noalign{\vskip 1.5mm}
\multicolumn{12}{c}{ImageNet16-120}\\
\noalign{\vskip 1.5mm}
\toprule
\noalign{\vskip 1.5mm} 
\multicolumn{12}{c}{Batch size}\\
\cline{3-4}\cline{5-6}\cline{7-8}\cline{9-10}\cline{11-12}
\multicolumn{2}{l}{ } & \multicolumn{2}{c}{2} & \multicolumn{2}{c}{4}           & \multicolumn{2}{c}{8}           & \multicolumn{2}{c}{16}          & \multicolumn{2}{c}{32}         \\
  $N_{a}$     &    $N_{init}$  & Validation      & Test            & Validation     & Test           & Validation     & Test           & Validation     & Test           & Validation     & Test    \\
\hline

\multirow{2}{*}{10}  & $10$         & $31.40 \pm 8.43$ & $30.96 \pm 8.66$ & $33.02 \pm 7.71$ & $31.85 \pm 8.62$ & $33.73 \pm 7.58$ & $33.39 \pm 7.83$ & $34.29 \pm 6.87$ & $34.04 \pm 7.15$ & $32.78 \pm 7.51$ & $32.37 \pm 7.83$ \\
                    & $100$        & $31.97 \pm 7.78$ & $31.49 \pm 8.02$ & $37.46 \pm 6.53$ & $37.39 \pm 6.78$ & $37.96 \pm 6.08$ & $37.91 \pm 6.33$ & $37.74 \pm 6.89$ & $37.70 \pm 7.13$ & $35.72 \pm 9.34$ & $35.59 \pm 9.69$ \\
\hline
\multirow{2}{*}{25} & $10$        & $32.09 \pm 8.11$ & $31.66 \pm 8.39$ & $32.31 \pm 8.36$ & $31.93 \pm 8.66$ & $34.95 \pm 6.78$ & $34.71 \pm 7.03$ & $34.11 \pm 6.50$ & $33.84 \pm 6.78$ & $32.93 \pm 6.95$ & $32.53 \pm 7.23$ \\
                    & $100$         & $36.67 \pm 6.63$ & $36.59 \pm 6.93$ & $37.86 \pm 5.93$ & $37.82 \pm 6.21$ & $38.44 \pm 5.90$ & $38.47 \pm 6.19$ & $38.42 \pm 6.30$ & $38.42 \pm 6.51$ & $36.44 \pm 9.49$ & $36.37 \pm 9.84$ \\
\hline
\multirow{2}{*}{100} & $10$         & $31.97 \pm 7.78$ & $31.49 \pm 8.02$ & $32.66 \pm 8.48$ & $32.36 \pm 8.75$ & $34.13 \pm 6.89$ & $33.81 \pm 7.11$ & $34.99 \pm 5.39$ & $34.69 \pm 5.66$ & $32.97 \pm 5.94$ & $32.46 \pm 6.19$ \\
                    & $100       $ & $36.93 \pm 6.38$ & $36.89 \pm 6.64$ & $38.78 \pm 5.60$ & $38.72 \pm 5.85$ & $39.11 \pm 5.03$ & $39.17 \pm 5.23$ & $38.68 \pm 6.34$ & $38.76 \pm 6.62$ & $36.73 \pm 9.92$ & $36.73 \pm 10.23$ \\
\noalign{\vskip 3mm}
\multicolumn{12}{c}{Batch size}\\
\cline{3-4}\cline{5-6}\cline{7-8}\cline{9-10}
\multicolumn{2}{l}{ } & \multicolumn{2}{c}{64}            & \multicolumn{2}{c}{128}         & \multicolumn{2}{c}{256}         & \multicolumn{2}{c}{512} & \\  
$N_{a}$     &    $N_{init}$ & Validation      & Test            & Validation     & Test           & Validation     & Test           & Validation     & Test \\
\cline{1-2} \cline{3-10}

\multirow{2}{*}{10}  & $10$         & $33.33 \pm 7.40 $ & $33.02 \pm 7.71 $ & $32.54 \pm 8.11$ & $32.85 \pm 7.78$ & $34.02 \pm 7.91$ & $33.72 \pm 8.30$ & $33.42 \pm 7.61$ & $33.11 \pm 7.94$  & \\
                     & $100$        & $36.55 \pm 8.16 $ & $36.47 \pm 8.49 $ & $36.21 \pm 8.12$ & $36.09 \pm 8.43$ & $35.64 \pm 8.42$ & $35.50 \pm 8.77$ & $35.04 \pm 8.40$ & $34.90 \pm 8.72$  & \\
\cline{1-2} \cline{3-10}
\multirow{2}{*}{25}  & $10$         & $33.39 \pm 6.87 $ & $33.02 \pm 7.11 $ & $32.32 \pm 7.95$ & $32.66 \pm 7.65$ & $33.90 \pm 7.74$ & $33.58 \pm 8.08$ & $33.37 \pm 7.65$ & $33.06 \pm 7.99$ & \\
                     & $100$        & $37.77 \pm 7.76 $ & $37.75 \pm 8.03 $ & $35.64 \pm 8.82$ & $35.51 \pm 9.13$ & $36.75 \pm 7.84$ & $36.71 \pm 8.14$ & $34.92 \pm 8.18$ & $34.74 \pm 8.45$ & \\
\cline{1-2} \cline{3-10}
\multirow{2}{*}{100} & $10$         & $34.02 \pm 6.41 $ & $33.64 \pm 6.64 $ & $32.69 \pm 7.06$ & $33.06 \pm 6.81$ & $34.08 \pm 7.39$ & $33.79 \pm 7.73$ & $34.04 \pm 7.12$ & $33.79 \pm 7.49$ & \\
                     & $100$        & $37.48 \pm 8.54 $ & $37.49 \pm 8.83 $ & $37.35 \pm 7.69$ & $37.27 \pm 8.05$ & $38.11 \pm 6.75$ & $38.11 \pm 7.06$ & $35.61 \pm 7.81$ & $35.46 \pm 8.08$ & \\  
\end{tabular}
}
\label{tab:overalltab}
\end{table}

\clearpage
%CIFAR100, architectures
\begin{figure}[H]
    \centering
    \begin{minipage}{.3\linewidth}
        \begin{subfigure}{\linewidth}
            \includegraphics[width=\textwidth]{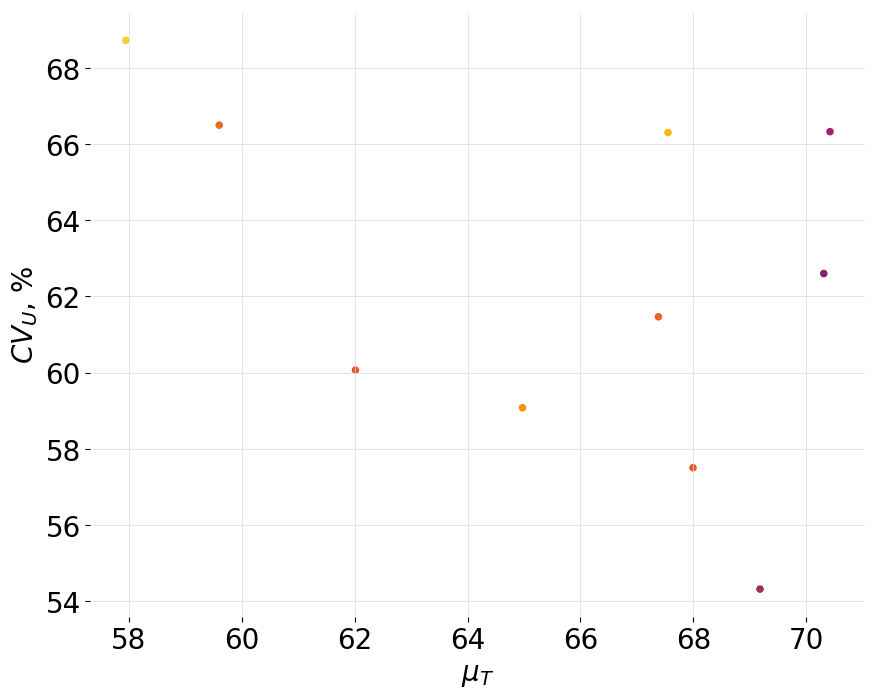}
            \caption{$N_{a}=10$}
        \end{subfigure} \\
        \begin{subfigure}{\linewidth}
            \includegraphics[width=\textwidth]{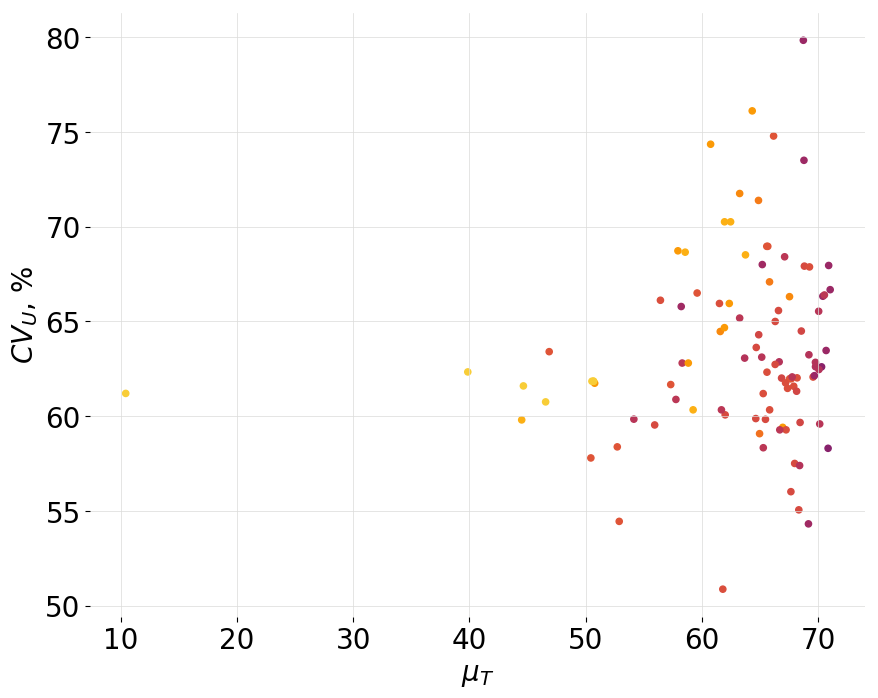}
            \caption{$N_{a}=100$}
        \end{subfigure} 
    \end{minipage}
    \begin{minipage}{.3\linewidth}
        \begin{subfigure}{\linewidth}
            \includegraphics[width=\textwidth]{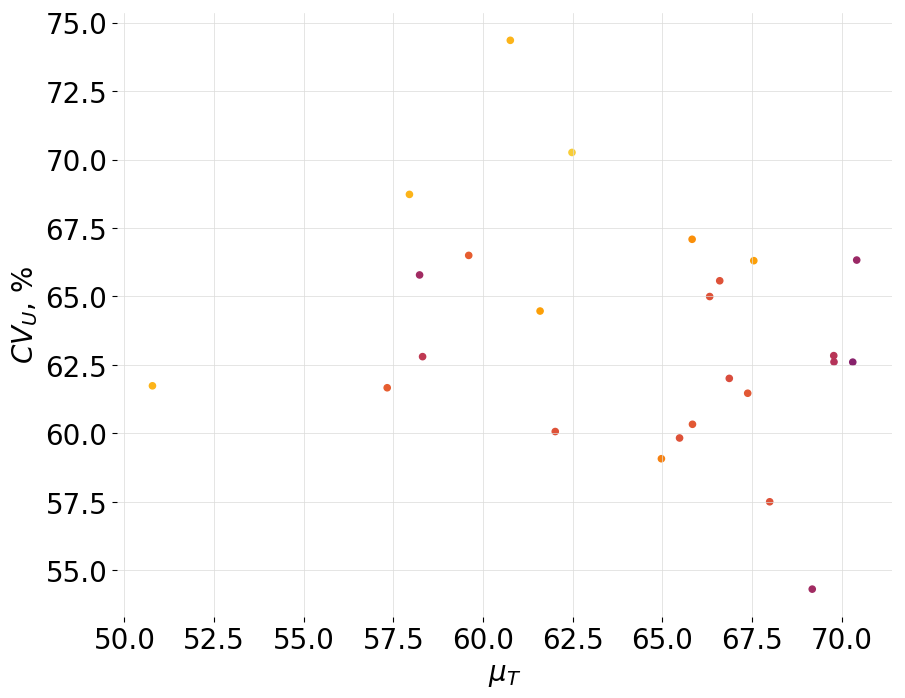}
            \caption{$N_{a}=25$}
        \end{subfigure} \\
        \begin{subfigure}{\linewidth}
            \includegraphics[width=\textwidth]{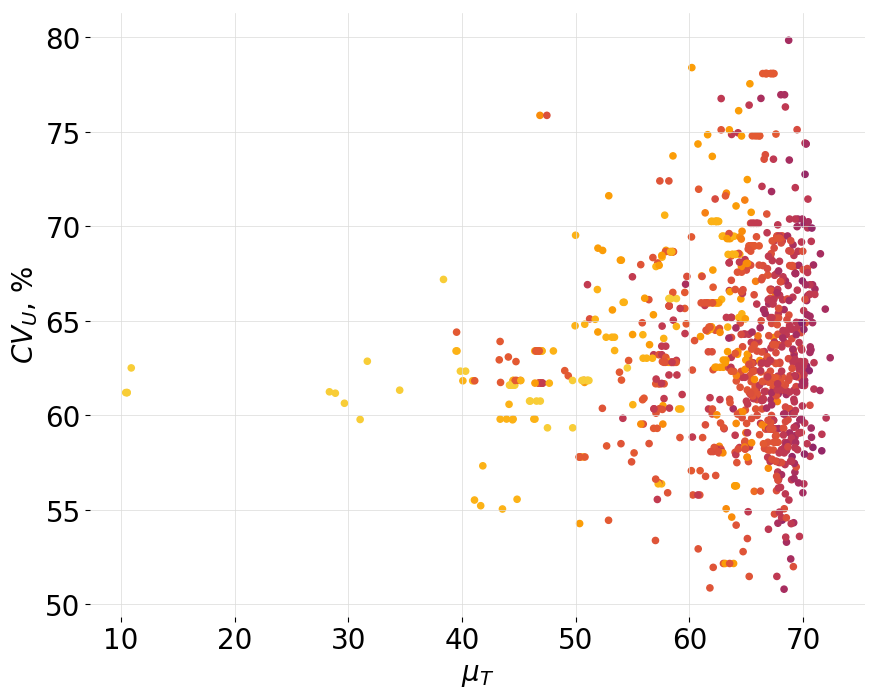}
            \caption{$N_{a}=1000$}
        \end{subfigure} 
    \end{minipage}
    \begin{minipage}{.3\linewidth}
        \begin{subfigure}{\linewidth}
            \includegraphics[width=\textwidth]{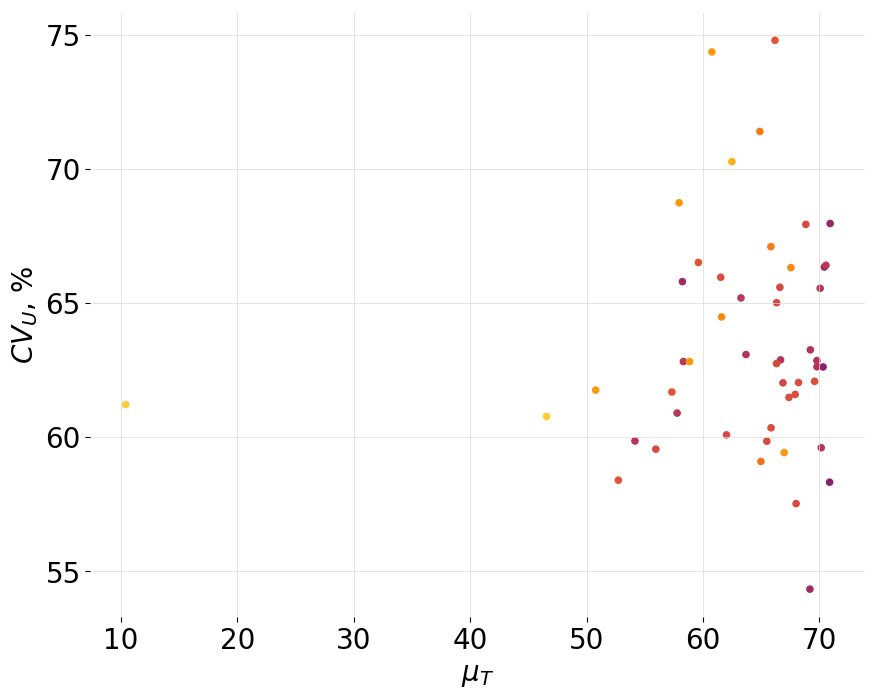}
            \caption{$N_{a}=50$}
        \end{subfigure} \\
        \begin{subfigure}{\linewidth}
            \includegraphics[width=\textwidth]{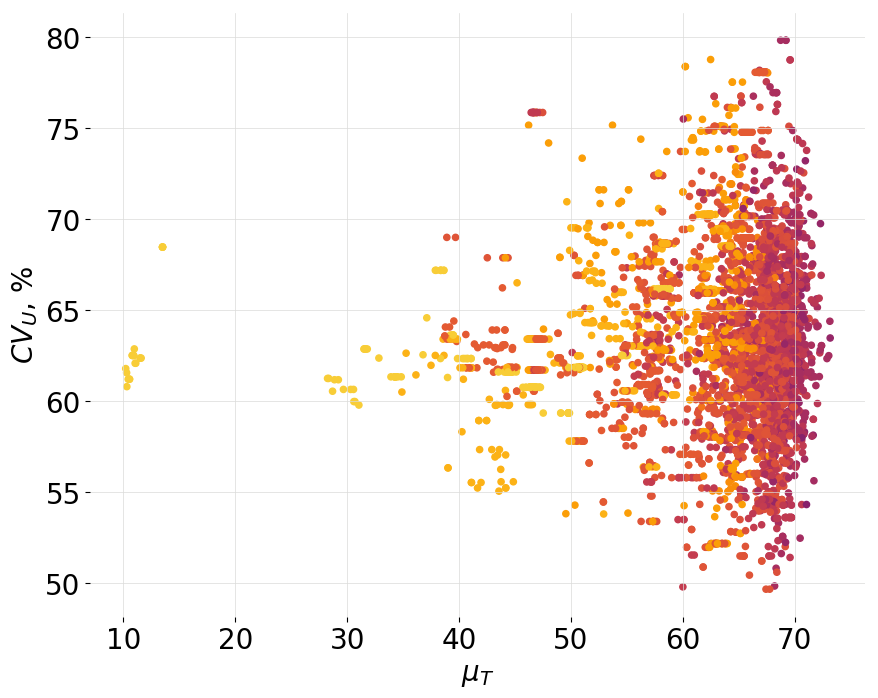}
            \caption{$N_{a}=5000$}
        \end{subfigure} 
    \end{minipage}
    \begin{minipage}{.07\linewidth}
            \begin{subfigure}[t]{\linewidth}
                \includegraphics[width=\textwidth]{figures/ColorBar.png}
            \end{subfigure}
        \end{minipage}
\caption{Comparison of the relative standard deviation $CV_{U}$ ($\rm\%$) performance against mean trained accuracy $\mu_{T}$ for CIFAR-100 \cite{krizhevsky2009learning} dataset for different number of selected architectures $N_{a}\in[10, 25, 50, 100, 1000, 5000]$. Statistics are computed over $N_{init}=100$ initialisations. One point represents one architecture. The colours represent the logarithm of the total number of trained parameters.} 
\label{fig:cifar100arch}
\end{figure}

% CIFAR100, initialisations
\begin{figure}[H]
    \centering
    \begin{minipage}{.3\linewidth}
        \begin{subfigure}{\linewidth}
            \includegraphics[width=\textwidth]{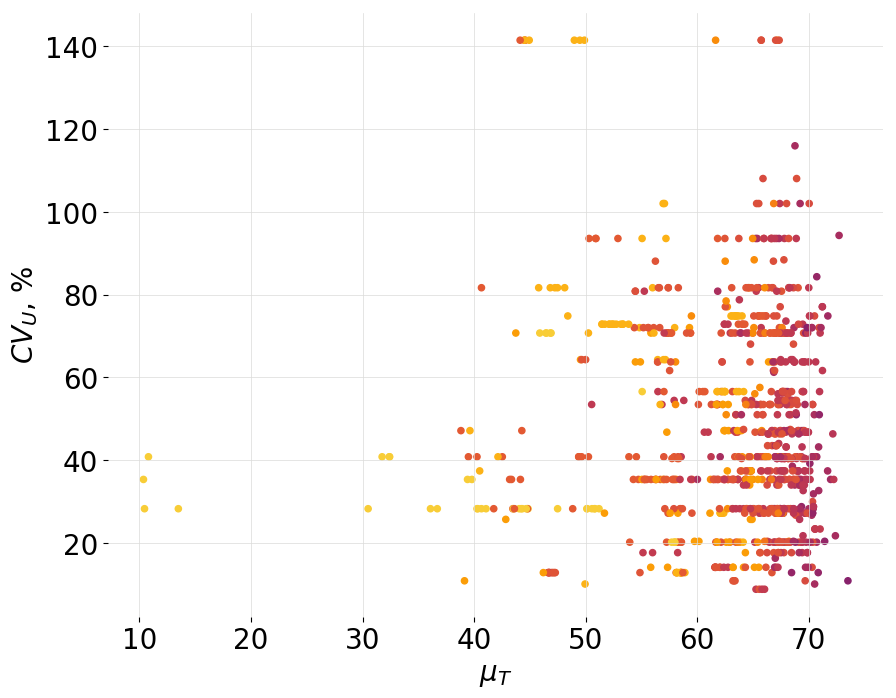}
            \caption{$N_{init}=3$}
        \end{subfigure} \\
        \begin{subfigure}{\linewidth}
            \includegraphics[width=\textwidth]{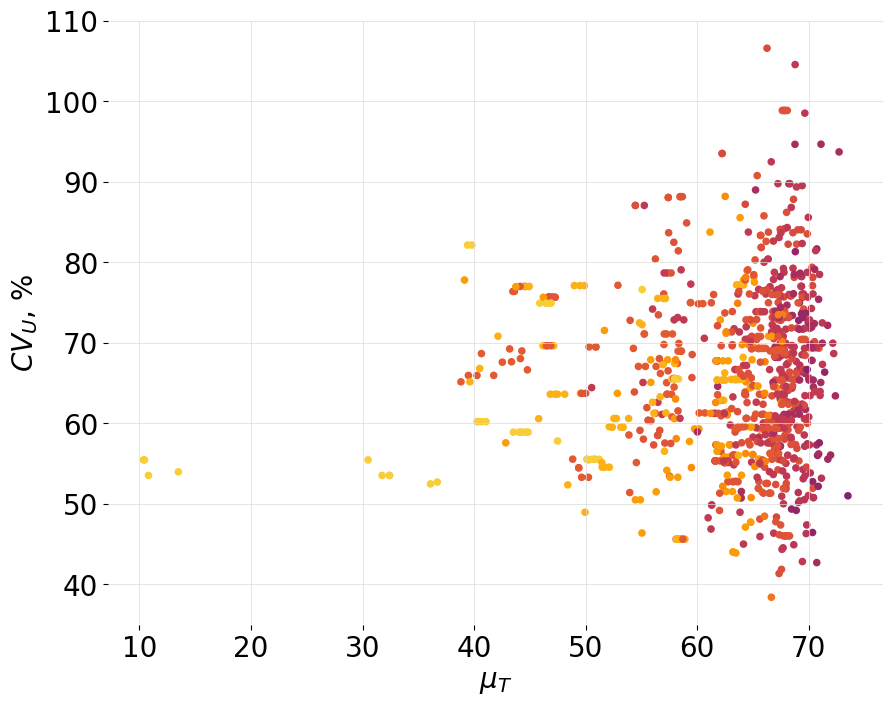}
            \caption{$N_{init}=25$}
        \end{subfigure} 
    \end{minipage}
    \begin{minipage}{.3\linewidth}
        \begin{subfigure}{\linewidth}
            \includegraphics[width=\textwidth]{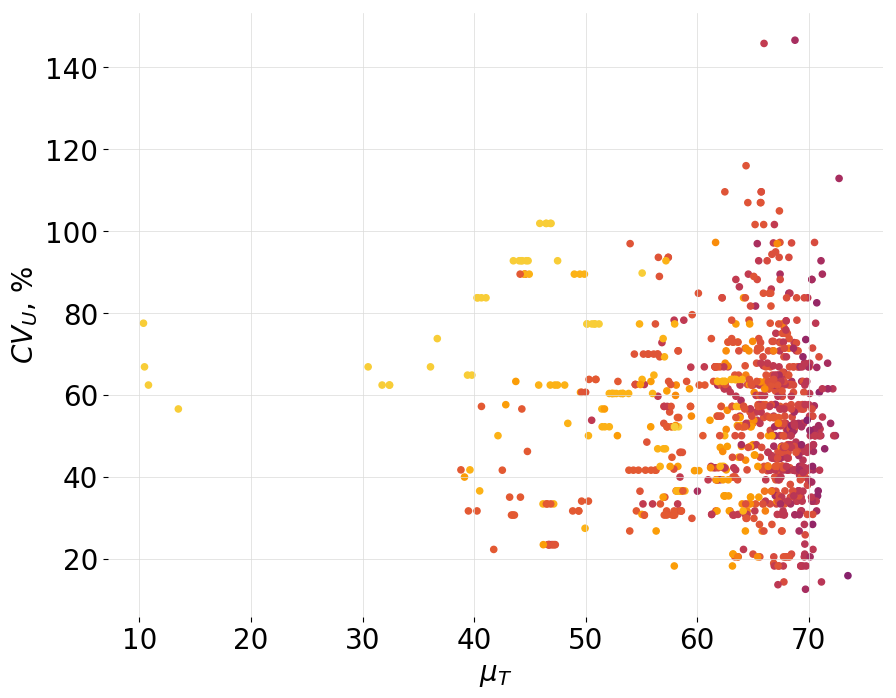}
            \caption{$N_{init}=5$}
        \end{subfigure} \\
        \begin{subfigure}{\linewidth}
            \includegraphics[width=\textwidth]{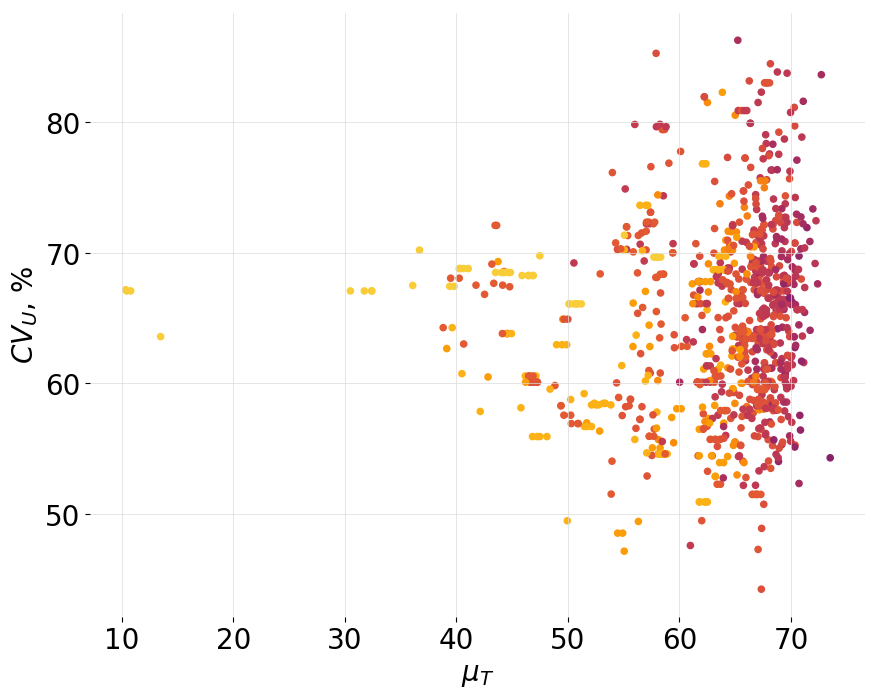}
            \caption{$N_{init}=50$}
        \end{subfigure} 
    \end{minipage}
    \begin{minipage}{.3\linewidth}
        \begin{subfigure}{\linewidth}
            \includegraphics[width=\textwidth]{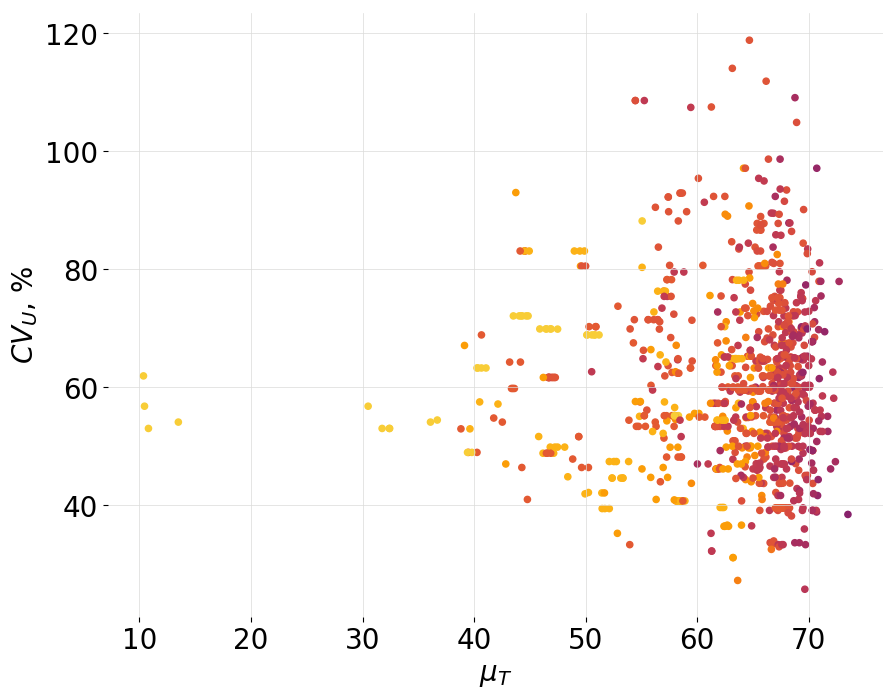}
            \caption{$N_{init}=10$}
        \end{subfigure} \\
        \begin{subfigure}{\linewidth}
            \includegraphics[width=\textwidth]{figures/cifar100/Correlaction_Score_vs_Accuracy_CIFAR100_256bs_100it_1000arch.png}
            \caption{$N_{init}=100$}
        \end{subfigure} 
    \end{minipage}
    \begin{minipage}{.07\linewidth}
            \begin{subfigure}[t]{\linewidth}
                \includegraphics[width=\textwidth]{figures/ColorBar.png}
            \end{subfigure}
        \end{minipage}
\caption{Comparison of the relative standard deviation $CV_{U}$ ($\rm\%$) performance against mean trained accuracy $\mu_{T}$ for CIFAR-100 \cite{krizhevsky2009learning} dataset. Statistics are computed over varying number of initialisations $N_{init}\in[3, 5, 10, 25, 50, 100]$.  One point stands for one architecture. The colours represent the logarithm of the total number of trained parameters.} 
\label{fig:cifar100init}
\end{figure}

% DONE
% IMAGENET16-120, architectures
\begin{figure}[H]
    \centering
    \begin{minipage}{.3\linewidth}
        \begin{subfigure}{\linewidth}
            \includegraphics[width=\textwidth]{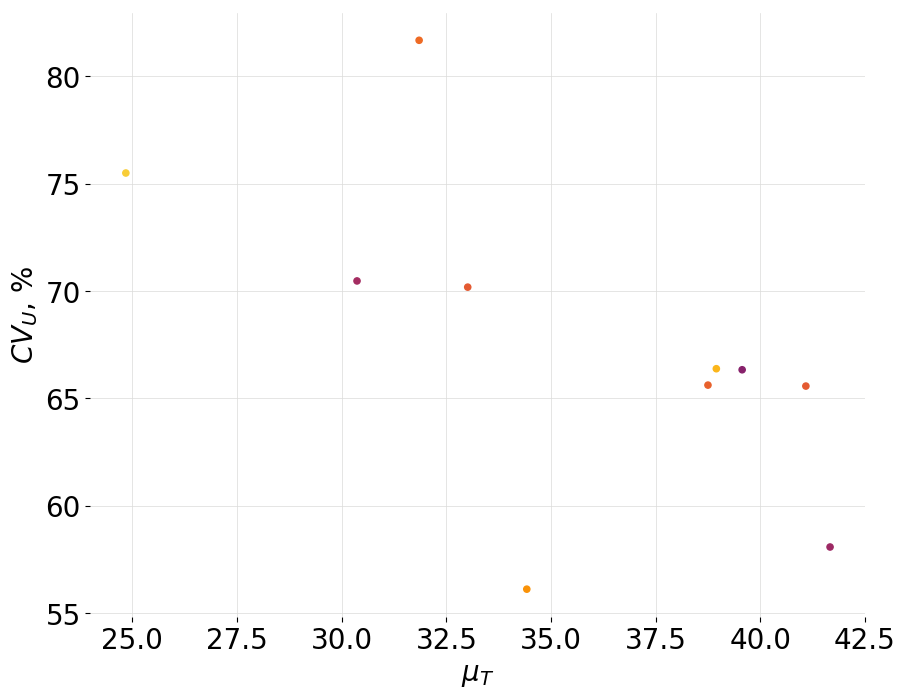}
            \caption{$N_{a}=10$}
        \end{subfigure} \\
        \begin{subfigure}{\linewidth}
            \includegraphics[width=\textwidth]{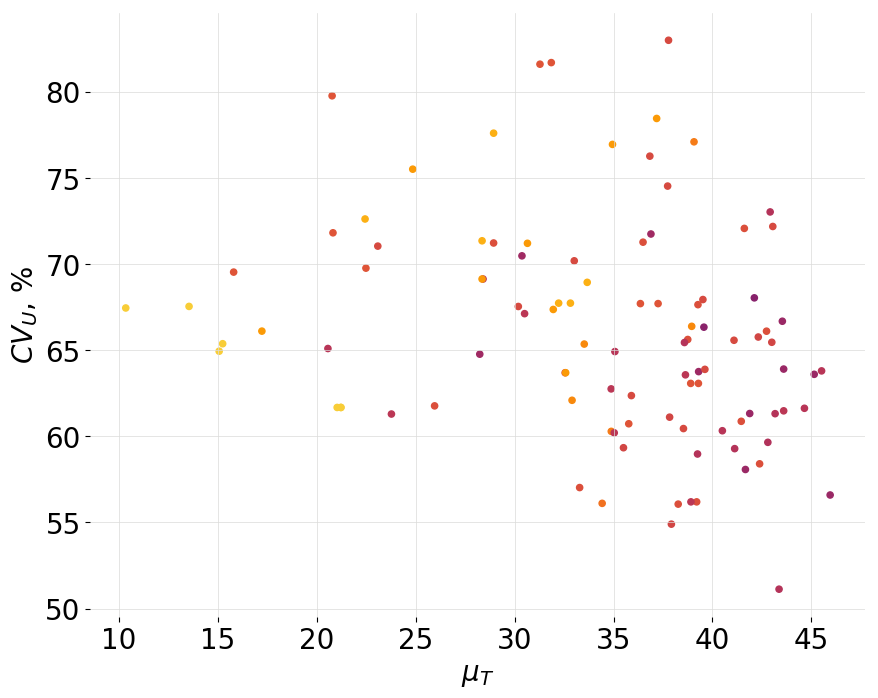}
            \caption{$N_{a}=100$}
        \end{subfigure} 
    \end{minipage}
    \begin{minipage}{.3\linewidth}
        \begin{subfigure}{\linewidth}
            \includegraphics[width=\textwidth]{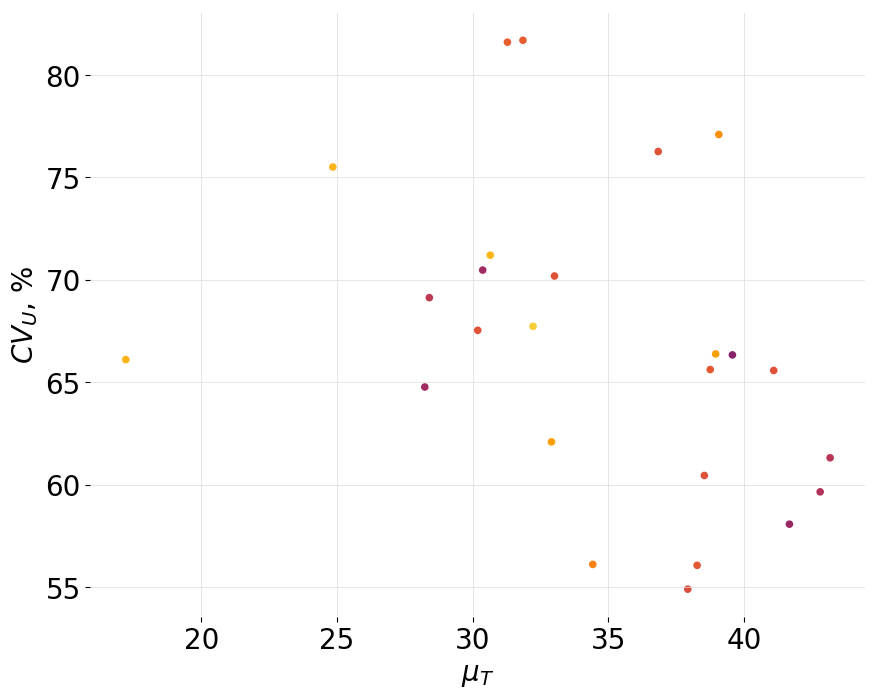}
            \caption{$N_{a}=25$}
        \end{subfigure} \\
        \begin{subfigure}{\linewidth}
            \includegraphics[width=\textwidth]{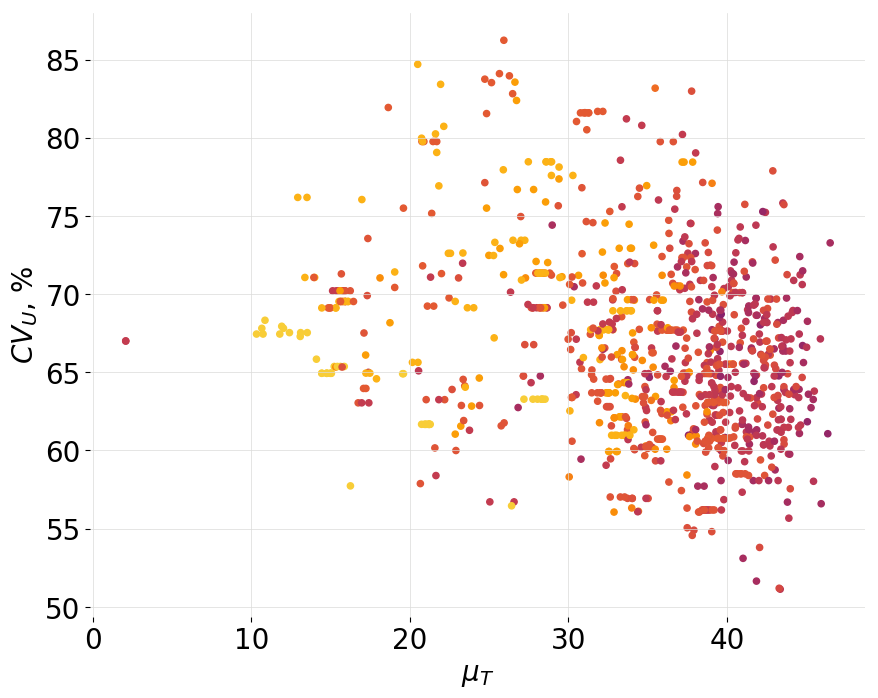}
            \caption{$N_{a}=1000$}
        \end{subfigure} 
    \end{minipage}
    \begin{minipage}{.3\linewidth}
        \begin{subfigure}{\linewidth}
            \includegraphics[width=\textwidth]{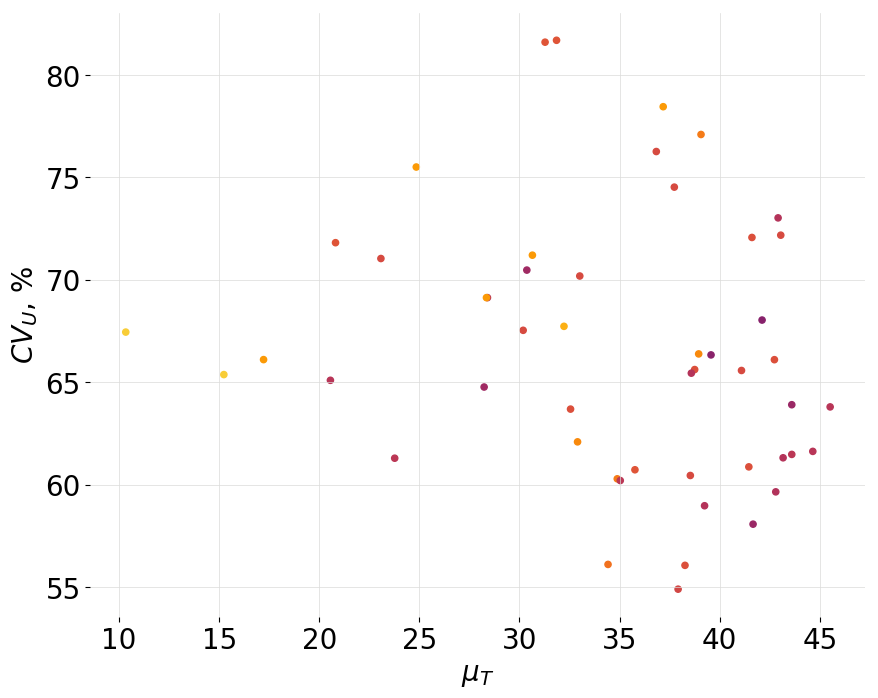}
            \caption{$N_{a}=50$}
        \end{subfigure} \\
        \begin{subfigure}{\linewidth}
            \includegraphics[width=\textwidth]{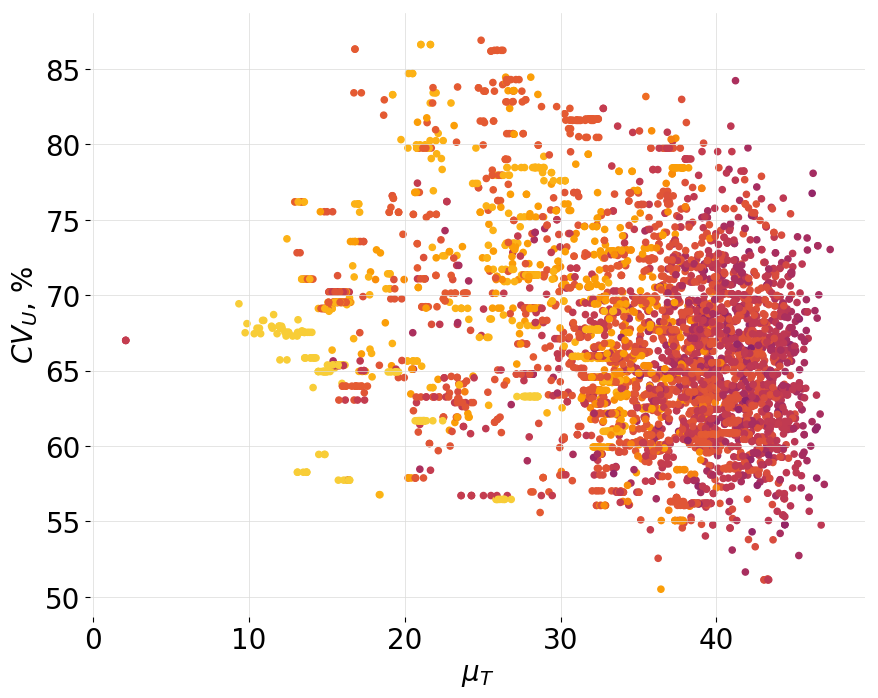}
            \caption{$N_{a}=5000$}
        \end{subfigure} 
    \end{minipage}
    \begin{minipage}{.07\linewidth}
            \begin{subfigure}[t]{\linewidth}
                \includegraphics[width=\textwidth]{figures/ColorBar.png}
            \end{subfigure}
        \end{minipage}
\caption{Comparison of the relative standard deviation $CV_{U}$ ($\rm\%$) performance against mean trained accuracy $\mu_{T}$ for ImageNet16-120 dataset \cite{chrabaszcz2017downsampled} for different number of selected architectures $N_{a}\in[10, 25, 50, 100, 1000, 5000]$. Statistics are computed over $N_{init}=100$ initialisations. One point represents one architecture. The colours represent the logarithm of the total number of trained parameters.} 
\label{fig:imagenet16120arch}
\end{figure}

% DONE
% IMAGENET16-120, initialisations
\begin{figure}[H]
    \centering
    \begin{minipage}{.3\linewidth}
        \begin{subfigure}{\linewidth}
            \includegraphics[width=\textwidth]{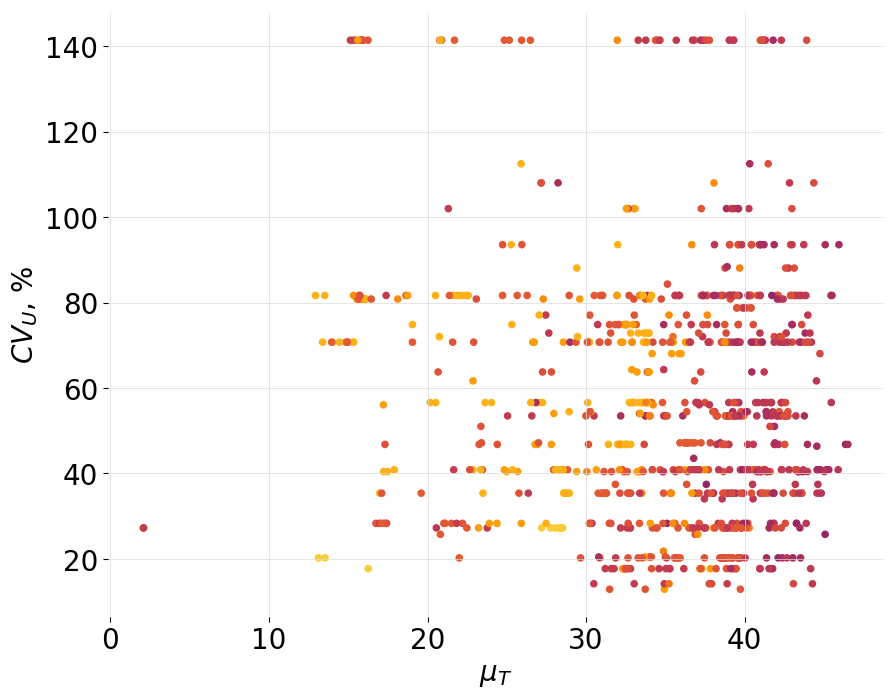}
            \caption{$N_{init}=3$}
        \end{subfigure} \\
        \begin{subfigure}{\linewidth}
            \includegraphics[width=\textwidth]{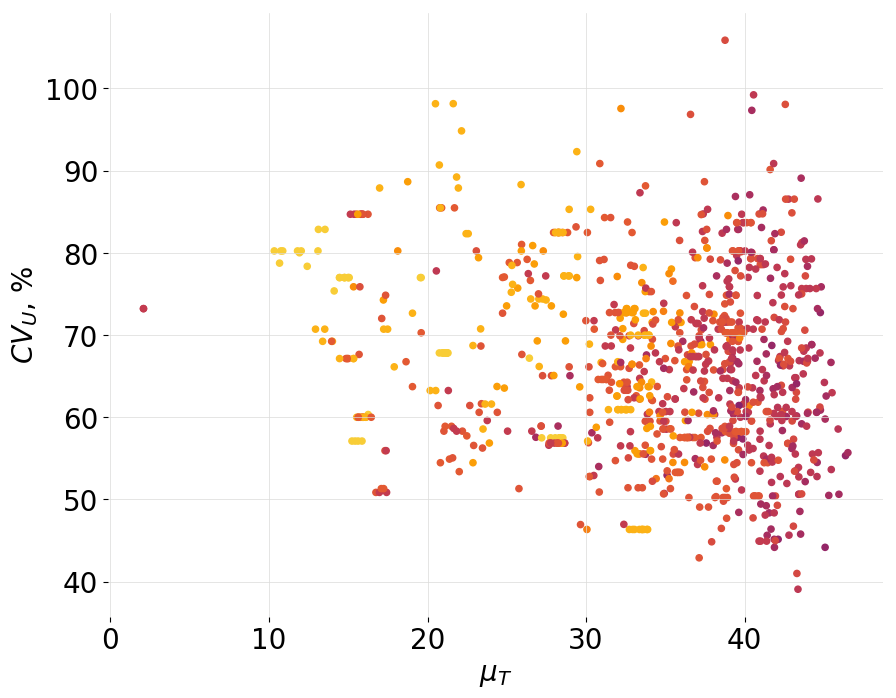}
            \caption{$N_{init}=25$}
        \end{subfigure} 
    \end{minipage}
    \begin{minipage}{.3\linewidth}
        \begin{subfigure}{\linewidth}
            \includegraphics[width=\textwidth]{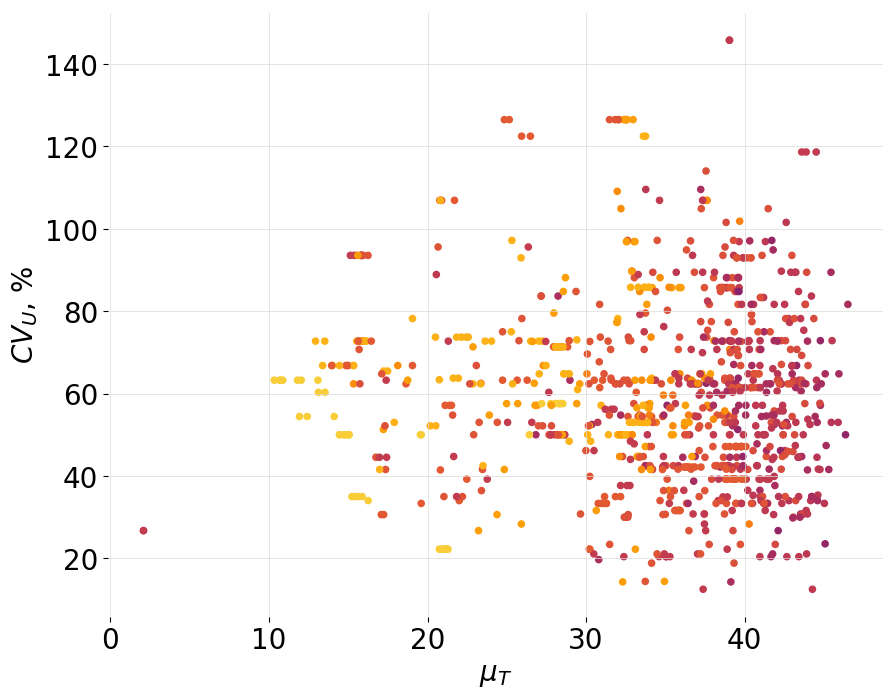}
            \caption{$N_{init}=5$}
        \end{subfigure} \\
        \begin{subfigure}{\linewidth}
            \includegraphics[width=\textwidth]{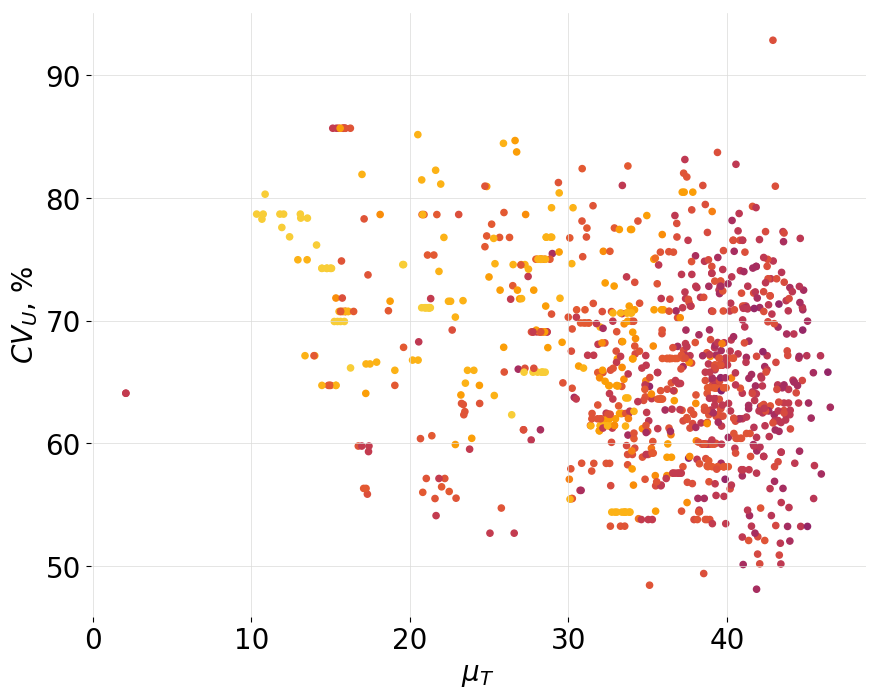}
            \caption{$N_{init}=50$}
        \end{subfigure} 
    \end{minipage}
    \begin{minipage}{.3\linewidth}
        \begin{subfigure}{\linewidth}
            \includegraphics[width=\textwidth]{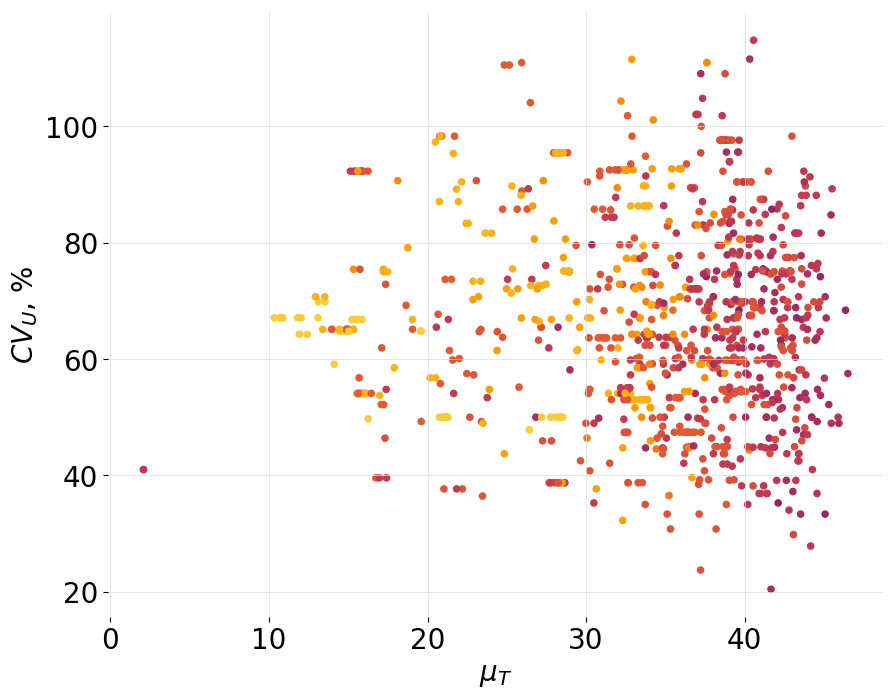}
            \caption{$N_{init}=10$}
        \end{subfigure} \\
        \begin{subfigure}{\linewidth}
            \includegraphics[width=\textwidth]{figures/imagenet/Correlaction_Score_vs_Accuracy_IMAGENET16-120_256bs_100it_1000arch.png}
            \caption{$N_{init}=100$}
        \end{subfigure} 
    \end{minipage}
    \begin{minipage}{.07\linewidth}
            \begin{subfigure}[t]{\linewidth}
                \includegraphics[width=\textwidth]{figures/ColorBar.png}
            \end{subfigure}
        \end{minipage}
\caption{Comparison of the relative standard deviation $CV_{U}$ ($\rm\%$) performance against mean trained accuracy $\mu_{T}$ for ImageNet16-120 dataset \cite{chrabaszcz2017downsampled}. Statistics are computed over varying number of initialisations $N_{init}\in[3, 5, 10, 25, 50, 100]$.  One point stands for one architecture. The colours represent the logarithm of the total number of trained parameters.} 
\label{fig:imagenet16120init}
\end{figure}

\end{document}